\documentclass[pdflatex,sn-basic]{sn-jnl}

\usepackage{amsmath,amssymb}
\usepackage{mathtools}
\usepackage{multirow}
\usepackage{float}
\usepackage[dvipsnames]{xcolor}
\usepackage{algorithmic}
\usepackage{booktabs}
\usepackage{tabularx}
\usepackage{siunitx}
\usepackage{graphicx}
\usepackage{subcaption}
\usepackage{makecell}
\usepackage{lmodern}

\newcommand{\textdef}[1]{\textit{#1}}
\newcommand{\citeposs}[1]{\citeauthor{#1}'s~(\citeyear{#1})}
\def\det{\textnormal{[\textsc{det}]}}
\def\W{\textnormal{[\textsc{w}]}}
\def\V{\textnormal{[\textsc{v}]}}

\usepackage[colorinlistoftodos]{todonotes}
\usepackage[dvipsnames]{xcolor}
\newcommand{\newcontents}[1]{#1}

\begin{document}

\title[Article Title]{A Comprehensive Evaluation of Semantic Relation Knowledge of Pretrained Language Models and Humans}

\author*[1]{\fnm{Zhihan} \sur{Cao}}\email{cao.z.ab@m.titech.ac.jp}
\author[1]{\fnm{Hiroaki} \sur{Yamada}}\email{yamada@c.titech.ac.jp}
\author[1,2]{\fnm{Simone} \sur{Teufel}}\email{simone.teufel@cam.ac.uk}
\author[1]{\fnm{Takenobu} \sur{Tokunaga}}\email{take@c.titech.ac.jp}

\affil[1]{\orgdiv{School of Computing}, \orgname{Institute of Science Tokyo}}
\affil[2]{\orgdiv{Department of Computer Science and Technology}, \orgname{University of Cambridge}}

\abstract{
\newcontents{Recently, much work has concerned itself with the enigma of what exactly pretrained language models~(PLMs) learn about different aspects of language, and how they learn it.
One stream of this type of research investigates the knowledge that PLMs have about semantic relations.
However, many aspects of semantic relations were left unexplored.
Generally, only one relation has been considered, namely hypernymy. 
Furthermore, previous work did not measure humans' performance on the same task as that performed by the PLMs. 
This means that at this point in time, there is only an incomplete view of the extent of these models' semantic relation knowledge.
To address this gap, we introduce a comprehensive evaluation framework covering five relations beyond hypernymy, namely hyponymy, holonymy, meronymy, antonymy, and synonymy.
We use five metrics (two newly introduced here) for recently untreated aspects of semantic relation knowledge, namely soundness, completeness, symmetry, prototypicality, and distinguishability.
Using these, we can fairly compare humans and models on the same task.
Our extensive experiments involve six PLMs, four masked and two causal language models. 
The results reveal a significant knowledge gap between humans and models for all semantic relations.
In general, causal language models, despite their wide use, do not always perform significantly better than masked language models.
Antonymy is the outlier relation where all models perform reasonably well. 
The evaluation materials can be found at \url{https://github.com/hancules/ProbeResponses}.
}
}

\maketitle

\section{Introduction}
\label{sec:intro}

What do pretrained language models~(PLMs) learn about human language?
This question has recently been a central topic of discussion in Natural Language Processing~(NLP) and Computational Linguistics~(CL).
PLMs are utilized in various situations but are not thoroughly understood. 
While initial work explored syntactic and factual knowledge~\citep{Petroni_2019, Rogers_2020, Cao_2021, Li_2022, Mruthyunjaya_2023}, more recently there are a number of studies focusing on lexical semantic knowledge, particularly knowledge about semantic relations~\citep{Ettinger_2020, Ravichander_2020, Hanna_2021}.

Semantic relations describe how the senses of two lexical items are related.
They are an important aspect of linguistic knowledge because they structure the vocabulary of natural languages~\citep{Miller_1991a,semantic_priming,semantics}.
This makes them essential for both human language comprehension and production.
On the modeling side, semantic relations are crucial for tasks such as text simplification, paraphrasing, natural language inference, and discourse analysis, as has been shown experimentally~\citep{Tatu_2005, Madnani_2010, Glavas_2015, Alamillo_2023}.
Therefore, it is beneficial and necessary for PLMs to learn semantic relations well.

The present study tries to establish to what extent they are able to do so. 
We extend the existing methodology by introducing a new evaluation framework.
We cover six relations, namely hypernymy, hyponymy, holonymy, meronymy, antonymy, and synonymy.
Our framework is also the first to shine a light on previously understudied properties of semantic relations and meta-relations.
Two kinds of comparisons are considered.
First, we compare models against humans on the same task, so that we can quantify the difference with the theoretically achievable ceiling.
Second, we compare two families of models, which differ in the pretraining tasks used.
Within families, we consider different sizes.
These comparisons allow us to identify which factors facilitate the acquisition of semantic relation knowledge by the models.
In sum, our evaluation framework  deeply explores both well-studied and previously unexplored properties, and it uses new metrics in a new comparative experimental setting.
By doing so, it adds both depth and width to the current knowledge about the quality of semantic relation knowledge that can be acquired by today's PLMs.

\section{Related Work}
\label{sec:literature}

\subsection{Probing for Hypernymy}
\label{sec:probing}
\textdef{Hypernymy} is a typical semantic relation. 
In hypernymy, one word (the hyponym) refers to a specific concept and another word (the hypernym) refers to a more general concept encompassing the hyponym's meaning.
For example, ``bird'' is a hypernym of ``robin''.
\textdef{Hyponomy} is the name of the opposite relation: ``robin'' is a hyponym of ``bird''.

In order to study the linguistic knowledge of PLMs, including hypernymy, several methods have been proposed in the past. 
Standard approaches include the use of probing classifers~\citep{Hewitt_2019,Hewitt_2020,Hall_2020,Madsen_2021,Belinkov_2022} and prompt-based probing~\citep{Petroni_2019,Ettinger_2020, Rogers_2020, Cao_2021, Li_2022}. 

A probing classifier is a neural classifier that takes a word embedding or a combination of word embeddings as input and determines whether a linguistic property of interest holds.
In the context of probing for hypernymy knowledge, a probing classifier may take the concatenation of the embeddings of \text{``robin''} and \text{``bird''} as input and perform a binary classification, determining if hypernymy holds.
The performance of the classifier can then be interpreted as the extent to which the linguistic property in question is successfully encoded in the word embeddings.
Probing classifiers have the disadvantage that they require training, and that they also introduce new parameters.
They can therefore encounter the problem of double interpretation, where the researcher needs to interpret two sets of parameters at the same time: one set coming from the pretrained model probed, and another from the probing classifier.
Such situations lead to circularity.

In contrast, \citet{Ettinger_2020} pioneered the study of hypernymy with prompt-based probing. 
In prompt-based probing, responses from a model are elicited using a \textit{prompt}, a textual string with slots, all of which are unfilled.
An example prompt for hypernymy is ``a  {\W}  is a  {\V}'' where {\W} is a slot for the word given (called the \textdef{target word} here), whereas {\V} will be predicted by a model.
As the second word, which is called \textdef{relatum}, depends on the target word and the relation described.
A \textit{probe} is a prompt where the target word has been filled in. 
For example, Ettinger used the string ``a robin is a {\V}'' as a probe. 
If a PLM has learned hypernymy well, it should be able to predict hypernyms of ``robin'' for {\V}, such as ``bird''.

Prompt-based probing does not introduce new parameters and so avoids the aforementioned circularity in interpretation.
Moreover, since prompt-based probing is a language modeling task, it aligns well with the pretraining task of PLMs. 
Therefore, we consider prompt-based probing the natural choice for exploring semantic relation knowledge.

Ettinger tested two models, BERT-base and BERT-large~\citep{bert}.
The models' responses were evaluated by comparison with correct answers.
In Ettinger's hypernym prediction setting, both the target word (a hyponym) and the correct answer (a hypernym) were restricted to nouns, which came from a psycholinguistic experiment conducted by \citet{Fischler_1983}.
Ettinger found that both BERT-base and BERT-large achieve an accuracy of around 0.40 and a $\mathrm{Precision@5}$ score of 1.00, but these numbers were based on only 18 target words. 

\citet{Ravichander_2020} performed similar experiments with a larger dataset, also using BERT. 
Some of the target words they originally wanted to use were sense-ambiguous, but they decided to remove these from the evaluation set, resulting in 576 unambiguous target words.
They also measured to which degree models are affected when target words are changed from singular to plural form, and found that BERT's accuracy dropped from 0.68 for singular target words to 0.44 for plural ones. 
\citeauthor{Ravichander_2020} concluded from this that BERT's hypernymy knowledge is not robust. 

\citet{Hanna_2021} further developed \citeauthor{Ettinger_2020}'s and \citeauthor{Ravichander_2020}'s prediction task, exploiting the fact that a semantic relation can be expressed by multiple prompts.
For the prompts, \citeauthor{Hanna_2021} adopted some lexico-syntactic patterns known from previous work ~\citep{Hearst_1992} to be effective at retrieving word pairs in hypernymy and hyponymy.
Such prompts include ``my favorite  {\W} is a  {\V}'', ``a  {\W}, such as a  {\V}'', ``a { {\W}} is a type of  {\V}'', ``a { {\W}} is a  {\V}''. 
\citeauthor{Hanna_2021} dealt with the problem of ambiguous target words in a different way from \citet{Ravichander_2020}. 
They attached an example sentence from SemCor~\citep{semcor} to each probe, whether the target word was ambiguous or not.
The example sentence was chosen in such a way that the target word was in a specific sense, namely the first WordNet sense whose hypernyms include the gold hypernym.
In \citeauthor{Hanna_2021}'s experiments, BERT reached an accuracy of 0.48 in the best setting. 
Contrary to intuition, the accuracy dropped by about 0.05 when the example sentences were used. 

Another modification of the original task is the evaluation data \citeauthor{Hanna_2021} used, which came from \textdef{category norms}.
Category norms are items that humans judge to be subtypes of a category given to them.
For instance, one category norm  for ``fish'' is ``tuna''; ``trout'' and ``salmon'' are others. 
\citet{Cohen_1957} were the first to collect category norms.
In their experiment, each participant wrote down category norms included in one of 43 categories given to them, as many as came to mind in 30 seconds.
\citeauthor{Cohen_1957}'s original category norm collection was later expanded by \citet{Battig_1969}, who added 13 new categories (for a total of 56 categories and 2,082 category-category norm pairs). 
Another collection effort by \citet{Exbattig} added 14 more new categories (for a total of 70 categories and 1,983 category-category norm pairs).

Most of the category norms stand in hypernymy relation with their category, but one can also find some rare cases where the category and the category norms are connected by meronymy, i.e., the part-of relationship.
For example, in \citeauthor{Exbattig}'s data, there is a category named ``part of building'', which contains the category norms ``window'', ``door'', and ``roof'', amongst others.

\citeauthor{Hanna_2021} applied the hypernym-based part of \citeauthor{Battig_1969}'s category norm data to their probing experiments in an obvious way, using category norms as target words and categories as gold relata.
They discarded categories and category norms that were tokenized into multiple tokens, ending up with a total of 863 norms paired with 25 categories.

\newcontents{Recently, large PLMs' hypernymy knowledge has also gained attention.
\citet{Tseng_2024} studied the semantic relation knowledge of Pythia~\citep{Pythia} and GPT-3~\citep{GPT-3}, using prompts consist of two parts: the gloss of the target word {\W} and the phrase indicating the relation explicitly, such as ``the hypernym of {\W} is {\V}''.
These prompts are meta-linguistic, which means that they could be applied to any relation.
This allows \citeauthor{Tseng_2024} to also treat verb hypernymy and noun holonymy.
Their metric is the path similarity between the predicted word and the correct answer.
The path similarity measures how close a pair of words is in the WordNet taxonomy, where 1 indicates that they are identical and 0 indicates no path exists between them.
They tested models on 702 prompts for noun hypernym, 583 for verb hypernym, and 164 for noun holonym.
The results showed that, for all models, the path similarities for noun hypernymy are generally higher than those for verb hypernymy and noun holonymy. 
For example, Pythia-12b achieved 0.51 for noun hypernymy but only 0.28 for verb hypernymy and 0.12 for noun holonymy.
}

\newcontents{
In summary, PLMs demonstrate a reasonable ability to predict hypernyms of target words.
However, several interesting avenues have been left underexplored.
The most obvious one is that there are several other established semantic relations apart from hypernymy. 
Although some studies have explored semantic relations other than hypernymy, such as antonymy by \citet{Hernandex_2024, Chanin_2024} and semantic relatedness by \citet{Santos_2024}, they are still few in number.
More importantly, they treated each relation individually.
However, as we will show, semantic relations never stand alone.
}

\subsection{Relations beyond Hypernymy}
\label{sec:metarel}
\newcontents{
Hypernymy is just one of the semantic relations, although it is an essential one. 
Other semantic relations have also been a long-standing research topic in psychological, theoretical, and computational linguistics.
Extensive prior work shows that the distinction among such semantic relations is non‑trivial for both humans and language models~\citep{Chaffin_1984, Chaffin_1990, Joosten_2010, Scheible_2013, Nguyen_2017, Ali_2019,Xie_2021}.
}

In psycholinguistics, \citet{Chaffin_1984} researched the similarities and differences between several semantic relations, as perceived by humans. 
The experiment used a semantic sorting task, where participants are instructed to group together~31 word pairs, each representing a particular semantic relation.
The relations came from five broad categories: contrast (including antonymy), similars~(including synonymy), class inclusion~(including hypernymy), part-whole~(holonymy), and case relations~(such as the agent-instrument relation and the agent-action relations, exemplified by ``farmer''/``tractor'' and ``dog''/``bark'').
The results showed that the human subjects were able to distinguish contrast (including antonymy) most easily from the other four relations. 
Similars (including synonymy) and class inclusion (including hypernymy) formed a second cluster, whereas case relations and part-whole formed a separate cluster each.

The similarities between hypernymy and holonymy have been extensively discussed in the semantic literature~\citep{Cruse_1986, Winston_1987, Joosten_2010}.
Both \citeauthor{Cruse_1986} and \citeauthor{Winston_1987} pointed out that hypernymy and holonymy are similar in that they both involve division and inclusion.
Word pairs related by hypernymy and holonymy always consist of a word referring to an entity that undergoes division, with the other word referring to the result of that division, whether as a part or a subclass.
\citeauthor{Joosten_2010} considered both relations under the term \textit{denotational inclusion}.
The similarity between the relations becomes even more obvious when collective nouns are involved. 
For example, we can say that a table is \textit{a kind of} furniture, and we can also say that it is \textit{a part of} furniture, expressions typically associated with hypernymy and holonymy~\citep{Joosten_2010}. 

Another distinction which is well-known to be difficult is that between antonymy and other relations such as synonymy and hypernymy.
For both the antonymy and synonymy relation, word pairs possess high \textdef{paradigmatic similarity}, i.e., the words in a pair are interchangeable.
Distributional methods, which are based on co-occurrence statistics, therefore struggle with the distinction between antonymy and synonymy~\citep{Mohammad_2013}.
The problem has motivated various sophisticated technical solutions \citep{Scheible_2013,Ono_2015,Glavas_2018, Wang_2021}. 

Antonymy and hypernymy are also difficult to distinguish for unsupervised distributional measures, as was shown experimentally by~\citet{Shwartz_2017}.
They used the hypernymy discrimination task, which consists of distinguishing word pairs that stand in hypernym relation, from word pairs in one other relation.
Each non-hypernymy relation was tested separately by creating a mixture of word pairs in it and in hypernymy relation. 
The non-hypernymy relations tested are antonymy, synonymy, meronymy, and the attribute relation\footnote{Attribute relation is the relation holding between an adjective and a related attribute, such as ``cold'' and ``temperature''.}. 
\citeauthor{Shwartz_2017} compared an extensive number of unsupervised distributional measures on this task, and found that in all experiment settings, it was always the antonymy mixture that yielded the lowest performance out of all mixtures. 

Synonymy and hypernymy are also closely related, as was confirmed during the creation of Hyperlex \citep{hyperlex}.
Hyperlex is a lexical resource of semantic relations (hypernymy, hyponymy, meronymy, synonymy, antonymy, \newcontents{co-hyponymy}\footnote{Co-hyponymy is the relation between two concepts that share a hypernym, such as ``hawk'' and ``robin'' whose hypernyms are "bird".}), holding between 2,616 word pairs of nouns or verbs\footnote{290 unrelated word pairs also exist in Hyperlex.}.
The original pairs in Hyperlex were sampled from WordNet~\citep{Miller_1995} and the University of Southern Florida Norms dataset~(USF)~\citep{USF}. 
Three human checkers were asked to verify whether the semantic relation holds for each pair sampled; the pair was discarded unless two of them agreed that it did. 
Next, different crowd workers were asked to assign a score to the remaining pairs, indicating the degree to which the pair satisfies hypernymy.
Note that for those pairs that were related in a non-hypernymy relation, the crowd workers should assign a low score. 
However, the humans' score for synonymy was close to that for hypernymy pairs if the two words were close to each other in the WordNet hierarchy (i.e., separated by at most two levels). 

\subsection{Symmetry and Prototypicality}
When assessing a model's knowledge about semantic relations holistically, it is necessary to consider not only if the model uses the semantic relations correctly, but also to what extent it learns what specific properties the semantic relations have.

\paragraph{Symmetry}
One such property is \textit{symmetry}, which is defined as follows: If a word pair $(w_1,w_2)$ is in a symmetric relation, then the reverse pair $(w_2,w_1)$ is also in the relation.

Symmetry is also a property of some factual knowledge relations.
For instance, the factual knowledge relation ``is a sibling of'' is symmetric. 
\citet{Mruthyunjaya_2023} proposed metrics in order to assess whether models learn properties of such factual relations, including symmetry. 
They used prompt-based probing and defined the concept of \textdef{reciprocal elicitation}: for any word pair (target word $w$ and relatum $v$) that forms a symmetric relation, the model should respond with the relatum, when given the target word, and also respond with the target word, when given the relatum.
For the probe ``Bart Simpson is a sibling of  {\V}'', they expected models to predict ``Lisa Simpson'', and for the converse probe ``Lisa Simpson is a sibling of {\V}'', to predict ``Bart Simpson''.

For every such word pair $(w,v)$, a prompt $p$, and a model $m$'s top $k$ items in the response $m_k(w,p)$, given $w$ and $p$, the symmetry score is the average of $\mu_k(w,v,p)$ over all $(w,v)$ and $p$.
\begin{align}
\label{eq:origin_sy}
    \mu_k(w,v,p;m) = \mathbb{I}[ v \in m_k(w,p) ] \times \mathbb{I}[ w \in m_k(v,p)]
\end{align}
where $\mathbb{I}[ P ]$ is the indicator function that becomes one when $P$ is true and zero otherwise. 
This score ranges between zero and one.

Intuitively, the symmetry score can be interpreted as the probability of reciprocal elicitation for a semantic relation $r$.
A high score means that agents were able to detect symmetry. 
\citeauthor{Mruthyunjaya_2023}'s results showed that, for symmetry, BERT outperforms even GPT-3~\citep{GPT-3}.

\paragraph{Prototypicality}
Another property of interest is \textbf{prototypicality}~\citep{Rosch_1973,Rosch_1975_1,Rosch_1975_2}.
\citet{Rosch_1975_1} posited that not all members of a category are equally exemplary of the category, but that there is a prototype, which is the best examplar among the members.
The prototypicality of any member of a category is then the degree to which it is exemplary of its category.
\citet{Rosch_1975_2} empirically found that, among the category norms of ``bird'' established by \citeauthor{Battig_1969},
``robin'' is the prototype, that ``penguin'' has the lowest prototypicality, and that ``raven'' and ``parrot'' are somewhere in the middle. 
There is a close relationship between prototypicality and hypernymy/hyponymy, as is implicit in the construction of her experiment\footnote{Despite this obvious relationship, Rosch did not explicitly use the term \textdef{hypernymy}.}.

There have been theoretical discussions of the prototypicality of \textdef{holonymy}. 
\citet{Taylor_1996} and \citet{Tversky_1990} presented top-down accounts, which emphasize that the whole~(holonym) is intrinsic in the conceptualization of the part~(meronym).
They therefore predict that the relationship between the whole and its mandatory parts should be tighter than between the whole and its optional parts. 
A building is a structure with walls, and walls are defined by their function within a building. 
This makes ``building'' a prototypical holonym of ``wall''.
But not all holonymy pairs are well described by these accounts, because some optional parts also play an important role. 
This is acknowledged in the bottom-up accounts~\citep[][both cited by \citet{Joosten_2010}]
{Lecolle_1998,Mihatsch_2000}, who state that a whole is formed by assembling a number of other individual wholes, each of which has a separate existence outside the holonymy relation.
For example, ``sky'' is a typical holonym of ``cloud'' (indeed, in our forthcoming experiments, it happens to be the holonym most frequently mentioned by humans). 
However, in sunny weather, the sky can be cloudless, so clouds are not intrinsic to the sky. 
This seems to make bottom-up theories descriptively more adequate, although neither proposes a prototype prediction mechanism.

The antonymy relation also shows prototypicality effects; this has been empirically confirmed with corpus experiments~\citep{Jones_2007} and human experiments \citep{Paradis_2009, Pastena_2016}.
The two other semantic relations of interest to us (meronymy and synonymy) have not been studied in connection with prototypicality, either theoretically and empirically. 
There are also no experimental studies that evaluate how neural models learn prototypicality for any relation.

\section{Methodological Considerations}
\label{sec:two issues}
The majority of the previous research studied hypernymy. 
Beyond hypernymy, there is also much theoretical and experimental knowledge about semantic relations.
Additionally, the literature has established several facts about meta-relations. 
However, when it comes to practical investigations of model behaviour, it is always only hypernymy that has been studied, even though it is merely one semantic relation among many.

Additionally, we have seen that the relevant theoretical literature has extensively studied meta-relations such as distinguishability between semantic relations (cf. section~\ref{sec:metarel}).
In stark contrast with this, the methodology previously used is unable to establish the degree to which a model mistakes one semantic relation for another, and methodologies for other meta-relations are non-existent. 
This leaves us with an incomplete understanding of the nature of semantic relations, and of the knowledge that PLMs have about semantic relations.

Therefore, we design new metrics for measuring prototypicality (a property of relations) and distinguishability (a meta-relation). 
Using these new metrics, and the established ones for symmetry, we study hyponymy, holonymy, meronymy, antonymy and synonymy, as well as hypernymy.
We further provide a direct comparison between models and humans on the same task.
Such a human ceiling will allow us to interpret the performance of models more meaningfully.

\paragraph{Word Senses}
Another recurrent problem for all probing experiments is that most target words from any source, are naturally sense-ambiguous words.
Previous experimental attempts to deal with sense ambiguity are suboptimal.
\citeauthor{Ravichander_2020} limited the target words they use to the unambiguous words.
This limits the research focus to an artificial subset of all possible words and relations, and has the practical disadvantage of considerably reducing the number of target words one can use.
The other existing solution is to provide a context of the target word sense, for example in the form of example sentences, as~\citeauthor{Hanna_2021} did, but this empirically harmed the performance of models. 
Our methodology offers an alternative solution to this problem.

\paragraph{Introduction of Relatum Sets}
Some previous experiments on relatum prediction assumed that there is only one correct relatum for each tuple.
\citet{Hanna_2021} acknowledge that ``orange'' has two gold relata: ``color'' and ``fruit''.
They, therefore, defined two separate gold tuples for this probe: (``orange'', HYP, ``color'') and (``orange'', HYP, ``fruit'').
However, when calculating the accuracy scores, they consider only the first item in the responses for both tuples, and then average over the two tuples.
In this setting, it is theoretically impossible for a model to achieve the full score (1) for any sense-ambiguous probe, even if the model had the ability to predict both relata.  
Whether the model predicts [``color'', ``fruit''] or [``fruit'', ``color''], the accuracy score is always 0.5\footnote{In the general case, the highest achievable accuracy score for target words with $n$ relata is $\frac{1}{n}$.}.

This means that the extent to which a model can predict \emph{all} relata of a target word is undervalued. 
The gold standard we want to define should treat target words with multiple relata more fairly.
We define gold standards as a set of relata, which we call \textdef{relatum set}. 
Working with relatum sets is particularly necessary when evaluating hyponymy and meronymy, since in these two relations multiple relata cases are likely to particularly frequent. 
Under the use of multiple relata, accuracy alone is no longer suitable for evaluation; instead, metrics borrowed from information retrieval are required.

The introduction of relatum sets has an important side-effect in that it enables the evaluation of models' recognition of prototypicality.
However, we do not know a priori if prototypicality holds for all relations of interest. 
This question needs to be experimentally established. 
Once the prototypicality of a relation is confirmed, the relatum set allows us to measure the degree to which the model has captured prototypicality, by a comparison to the human responses. 

\paragraph{Determiners}
A confounder in the interpretation of numerical results is the use of definite and indefinite determiners in the probes. 
Previous researchers routinely used probes that include indefinite determiners, such as \citeposs{Ettinger_2020} ``a robin is a {\V}''.
The English indefinite determiner changes its form from ``a'' to ``an'' if the following word's pronunciation starts with a vowel. 
Choosing ``a'' or ``an'' in a probe before the {\V} slot would therefore bias the prediction by models towards relata with an initial vowel or consonant.

\citeauthor{Ettinger_2020} ran experiments using both types of determiners, comparing pairs of probes differing only in the determiners used.
Manual inspection of the responses showed that BERT indeed always adhered to the morphophonetic rule.
For example, given the probe ``a hammer is a {\V}'', BERT-large predicts [``hammer'', ``tool'', ``weapon'', ``nail'', ``device''], whereas given the probe ``a hammer is an {\V}'', it predicts [``object'', ``instrument'', ``axe'', ``implement'', ``explosive''].
This suggests that the BERT models were able to utilize the grammatical information contained in the determiner as a clue. 

\citeauthor{Ravichander_2020}, when faced the problem of which indefinite determiner to place before {\V}, chose to always use the determiner that morphophonetically fits with the gold standard answer.
For example, the determiner in the probe ``a moth is an {\V}'' was chosen to be ``an'' exactly because the gold standard answer was ``insect''.
However, when the determiner acts as a clue for the model, it becomes impossible to disentangle how much of the results is due to the model's semantic knowledge and how much to the clue. 
Choosing randomly also does not solve the problem. 
Our solution uses both probes with ``a'' and probes with ``an'' and merges the results in a statistical manner.

\paragraph{Model Comparison}
Another understudied aspect is the comparison between different types of PLM models on the semantic probing task.
BERT, one of the models widely studied so far, is a masked language model~(MLM). 
MLMs are pretrained on the masked language modeling task, in which a model is asked to recover tokens in a given sentence that are randomly masked.
But recently, causal language models~(CLMs) such as Llama~\citep{llama, LLaMa3} have shown high performance in many tasks and thus gained attention.
CLMs are pretrained on next-token prediction, the task of predicting the rightmost word given a sequence of words.
The difference in pretraining tasks means that the models make their decisions based on different kinds of information. 
MLMs consider the context of both sides of the masked word, while CLMs consider only the preceding context.
Previous studies in a number of tasks found a large influence of the type of context used in PLM pretraining on performance. 
For instance, for factual knowledge, MLMs are superior over CLMs~\citep{Petroni_2019, Cao_2022, Mruthyunjaya_2023}.
We are the first to study the role of bidirectional contexts in the recognition of semantic relations. 
To be fair to both MLMs and CLMs, we have designed all our prompts in such a way that they end with the slot~{\V}. 

\paragraph{Model Size}
Apart from model type, model size may also matter.
For pretraining tasks, \citet{Kaplan_2020} showed that if the corpus size is fixed, larger models show smaller losses and thus better performance. 
This regularity was found to be empirically valid for other sentence completion tasks~\citep{GPT-3}, not only for pretraining tasks.
However, there are contrary reports from factual relation recognition that smaller models~(BERT and RoBERTa) outperform larger models~(GPT-4 and GPT-3) in the determination of some properties~\citep{Mruthyunjaya_2023}. 
When it comes to semantic relation tasks in general, it is unknown which of these tendencies is stronger.

We always first establish human performance for each task and then use it as the measuring stick for models' performance. 

The rest of the article is structured as follows.
Section~\ref{sec:method} describes the material collection.
We will then describe our proposed evaluation metrics in Section~\ref{sec:metrics}.
The following Sections~\ref{sec:human_exp} and \ref{sec:model_exp} will present the settings of human and model experiments, with results following in Section~\ref{sec:results}.

\section{Data}
\label{sec:method}

Our evaluation employs prompt-based probing.
In order to carry out the evaluation, we need prompts, target words and a gold-standard relatum set for each target word.
We will now explain how we collected them.

\subsection{Prompt Design}
\label{sec:prompt_design}
The underlying object we operate over is called a word-relation-relatum tuple~(tuple in short). We denote it by $t^r = (w, r, v)$, where $r\in R$ is a semantic relation, $w$ is a target word and $v$ is an $r$-relatum, i.e., a word standing in relation $r$ to $w$\footnote{In what follows, we omit the $r$- in the term if it is clear which specific relation is meant.}.
$T^r$ is the set of such tuples.
Our set of relations $R$ consists of hypernymy~(HYP), hyponymy~(HPO), holonymy~(HOL), meronymy~(MER), antonymy~(ANT), and synonymy~(SYN).

For each relation $r$, we construct a set of prompts $p^r\in P^r$.
We reuse \citeauthor{Hanna_2021}'s prompts for hypernymy.
For the other relations, we hand-craft new prompts.
In total, we use seven prompts for hypernymy, synonymy and holonymy; four prompts for hyponymy; six prompts for meronymy; and nine prompts for antonymy\footnote{A full list of prompts can be found in Appendix~\ref{app:prompts}.}.

Examples follow. 
\begin{equation}
\begin{aligned}
    p^\mathrm{HYP} &= \text{`` {\det}  {\W} is a kind of  {\det}  {\V}''},\\
    p^\mathrm{HPO} &= \text{``the word  {\W} has a more general meaning than the word  {\V}"},\\
    p^\mathrm{HOL} &= \text{`` {\det}  {\W} is a part of  {\det}  {\V}"},\\
    p^\mathrm{MER} &= \text{`` {\det}  {\W} has  {\det}  {\V}"},\\
    p^\mathrm{ANT} &= \text{`` {\det}  {\W} is the opposite of  {\det}  {\V}"},\\
    p^\mathrm{SYN} &= \text{`` {\det}  {\W} is also known as  {\det}  {\V}"}.  
\end{aligned}
\end{equation}

Note that some prompts do not require any determiner, but others do. 
The notation {\det} expresses that either ``an'' or ``a'' is chosen, based on certain conditions to be discussed later.
Our prompts are formulated such that {\W} always precedes {\V}, and that there is no token after the {\V} slot.

\subsection{Target Words and Probes}

\paragraph{Tuples}
In order to obtain our set of tuples $T^r$, we use existing word-relation-relatum tuples from Hyperlex~\citep{hyperlex} and the category norm corpus by \citet{Exbattig}.
We use only those tuples where both $w$ and $v$ are nouns, and where both are contained in the intersection of the vocabularies of all models that will be tested in the experiment.
This resulted in a total of 1,340 tuples: 145 for hypernymy, 800 for hyponymy, 234 for meronymy, 52 for antonymy and 109 for synonymy.
Note that none of the sources contributed any holonymy tuples.

To get more tuples, we expand our set of tuples by symmetric augmentation.
Symmetric augmentation can be applied to symmetric relations (here: antonymy and synonymy) by adding tuples where $v$ and $w$ are swapped, as follows:
\begin{gather}
T^{r,aug} = T^r \cup \{(w,r,v) \mid \forall (v,r,w)\in T^r\}.
\end{gather}

Symmetric augmentation can also be applied to those relations that have a \textdef{reverse} relation.
If $r_1$ and $r_2$ are reverse relations of each other, the following holds:
\begin{gather}
    (w,r_1,v)\in T^{r_1} \iff (v,r_2,w)\in T^{r_2}.
\end{gather}
Hypernymy and hyponymy are reverse relations of each other; so are holonymy and meronymy.
A small change to the symmetric augmentation procedure is necessary.
For reverse relations $r_1$ and $r_2$, symmetric augmentation proceeds as follows:
\begin{gather}
    T^{{r_1},aug} = T^{r_1} \cup \{(v,r_1,w) \mid \forall (w,r_2,v) \in T^{r_2} \},\\
    T^{{r_2},aug} = T^{r_2} \cup \{(v,r_2,w) \mid \forall (w,r_1,v) \in T^{r_1} \}.
\end{gather}

For example, we can reverse the meronymy tuple (``building'', MER, ``wall'') to obtain a new holonymy tuple (``wall'', HOL, ``building'').
Note that if there are any duplicate tuples, they are removed to form the set $T^{r,aug}$.
We will simplify notation after augmentation and use $T^{r}$ to refer to $T^{{r},aug}$.
After symmetric augmentation, the total number of tuples has risen 1.78 fold (2,390, from 1,340).
For holonomy, this process creates the only tuples in existence (234 tuples). 
\newcontents{
For other relations, the number of added tuples is 658 for hypernymy, 109 or synonymy, 46 for antonymy and~3 for hyponymy.
}

\paragraph{Target words}
Target words can be extracted from tuples as follows.
For each relation~$r$, we form $W^r$ as the set of target words $w$ from all tuples in $T^r$. 
Except for duplicates, each tuple contributes a target word.
$W$, the union of $W^r$ for different relations $r$, denotes the set of unique target words in all experiments, independent of relation. 

\paragraph{Probes}
A probe $x^r \in X^r = \{\nu(w^r, p^r)\mid \forall w^r\in W^r, \forall p^r\in P^r\}$ is then a string created by applying the verbalization function $\nu$ to a target word $w^r$ and a prompt $p^r$.
The verbalization function $\nu$ always assigns $w^r$ to the first slot {\W} and leaves the second slot {\V} empty. 
An example is
\begin{align}
    w^{\mathrm{HOL}}&=\text{``wall''},\\
    p^\mathrm{HOL}&=\text{``{\det}  {\W} is a part of  {\det}  {\V}"},\\
    \nu (w^\mathrm{HOL}, p^\mathrm{HOL}) &= \text{``a wall is a part of  {\det}  {\V}''}.
\end{align}

For the determiner before the target word, the function selects the morphophonetically correct form, as is uncontroversial and commonly done in previous work. 
For the indefinite determiner before {\V}, more thought is required. 
We explain our treatment in Section~\ref{sec:determiner issue}.

The total number of probes we create is 10,507; for each relation, the component is the product of prompts and target words. 
Statistics for each relation can be gleaned from Table~\ref{tab:relation_statistics}\footnote{The total over target words reported in the table is the sum over $|W^r|$. Because some target words are associated with more than one relation, this sum is different from $W$, the total number of unique target words, which is 1,266.}.

\begin{table}[htpb]
    \centering 
    \caption{Statistics of prompts, target words, and probes after augmentation.}
    \begin{tabular}{l|rrr}
    \toprule
    Relation &  \makecell{$|P^r|$\\(prompts)} 
    &  \makecell{$|W^r|$\\(target words)}  
    &  \makecell{$|X^r|$\\(probes) } \\

    \midrule
    Hypernymy~(HYP)  & 7  & 687  & 4,809 \\
    Hyponymy~(HPO)   & 4  & 309  & 1,236 \\
    Holonymy~(HOL)   & 7  & 186  & 1,302 \\
    Meronymy~(MER)   & 6  & 144  &   864 \\
    Antonymy~(ANT)   & 9  &  91  &   819 \\
    Synonymy~(SYN)   & 7  & 211  & 1,477 \\
    \midrule
    TOTAL          & 40  & 1,261 & 10,507 \\
    \bottomrule
    \end{tabular}
    \label{tab:relation_statistics}
\end{table}

\subsection{Relatum Sets}
So far, we have constructed probes that we will give as inputs to models and as stimuli to humans.
We now create $r$-relatum sets $Y^r$ that we can use for evaluation, for each target word $w\in W$ and relation $r$. 
We start by considering which properties good gold standards for our task would have. 

First, we want a sufficient number of $r$-relata for each target word in $W$.
This is important for a fair evaluation of relatum prediction ability.
If a dataset has only few relata per target word, we are lacking information about what the potential relatum set could look like, resulting in sparse data bias.
We therefore need larger relatum sets.

Second, each target word should be associated with as many relations as possible.
One of the abilities that we are going to evaluate is the degree to which models and humans can distinguish relations from each other. 
In principle, the more relations are present, the better the resulting evaluation should be.
At a minimum, each target word needs to be associated with two relations to make this measurement possible; at a maximum, each relation can be confused with five other relations. 

Unfortunately, the current relatum sets do not fulfill these two criteria.
On average, they contain only~1.1~(antonymy) to~2.6~(hyponymy) relata per target word, and the average number of relations associated with each target word is only~1.5.
This means that a majority of target words are associated with only one relation, making it impossible to assess models' ability to distinguish between relations. 
Therefore, it is desirable to increase the number of relata per relatum set, as well as the number of associated relations for each target word.

For a given relation $r$ and a target word $w$, we increase the $r$-relata in the $r$-relatum set $Y^r$ as follows.
We first retrieve all possible word senses of $w$ in WordNet.
Then, for each word sense that has $r$-relata documented in WordNet, we update $Y^r$ by adding the new $r$-relatum, unless it is not included in the models' shared vocabulary.
In other words, all possible $r$-relata of any sense of the target word are included in the expanded $Y^r$.
For example,  \text{``ending/1''} is a synonym of \text{``termination/4''}, and  \text{``ending/3''}  is a synonym of \text{``conclusion/6''}. 
The resulting synonym set for \text{ending} therefore includes both \text{``termination''} and \text{``conclusion''}.
This is so despite the fact that they refer to different senses of  \text{``ending''}\footnote{Note that the construction of synonym set for \text{``termination''} or \text{``conclusion''} results in a different expanded synonym set.}.
We additionally include indirect hypernyms and hyponyms of either of the target word senses, defined as those which lie within a path length of two in the WordNet hierarchy. 

This procedure can result in a situation where more than one semantic relation holds between two word forms.
For example, WordNet lists \text{``conclusion/3''} as a hyponym of \text{``ending/3''} and at the same time lists \text{``conclusion/4''} as a synonym of \text{``ending/4''}.
The result is two relations holding between the word forms \text{``conclusion''} and \text{``ending''}\footnote{The problem arises because our evaluation is performed at the word form level (as is the common approach), and not the sense level. 
If we were able to evaluate with senses disambiguated, we would be able to leave these relata in, with added profit.}. 
We therefore solve this problem by removing all relationally ambiguous relata for each target word. 
After this step, we have a guarantee that for each target word, the relatum sets of different relations are mutually exclusive. 

\begin{table}[htpb]
    \centering 
    \caption{Sizes of relatum sets before and after expansion per relation. 
    }
    \begin{tabular}{l|cc}
    \toprule
    Relation & Before Expansion & After Expansion \\
    \midrule
    Hypernymy~(HYP) & 1.2 $\pm$ 0.5 &  8.9 $\pm$  7.1 \\
    Hyponymy~(HPO)  & 2.6 $\pm$ 4.0 & 20.5 $\pm$ 43.0 \\
    Holonymy~(HOL)  & 1.3 $\pm$ 0.6 &  2.5 $\pm$  2.1 \\
    Meronymy~(MER)  & 1.7 $\pm$ 3.5 &  2.9 $\pm$  4.4 \\
    Antonymy~(ANT)  & 1.1 $\pm$ 0.3 &  1.1 $\pm$  0.4 \\
    Synonymy~(SYN)  & 1.1 $\pm$ 0.2 &  3.0 $\pm$  2.7 \\
    \bottomrule
    \end{tabular}
    \label{tab:relatum_set}
\end{table}

The expansion increases the average relatum set sizes, as can be seen from Table~\ref{tab:relatum_set}.
The final average relatum set sizes range from 1.1 for antonymy to 20.5 for hyponymy. 
The average size of expanded hyponym sets is far larger than others because of the nature of hyponymy; as we descend the WordNet hierarchy to retrieve hyponyms, the number of hyponyms increases. 
The expansion also increases the average number of relations associated with each target word from 1.1 to 2.9.

These relatum sets constitute our gold standard in the upcoming evaluation.
Of course, both humans and models can respond with a word that is not in the $r$-relatum set for either relation $r$.
We call such words OOR~(out of the relatum set).

\section{Metrics}
\label{sec:metrics}
The proposed evaluation framework consists of five metrics, two of which are novel.
The novel metrics are called prototypicality and distinguishability. 
Prototypicality evaluates a property of semantic relations.
Distinguishability evaluates agents' ability to distinguish relations from each other.
Soundness and completeness measure the performance of relatum prediction under the multiple relata setting.
Symmetry has been studied before, but only with factual relations. 

To calculate all metrics, we need a ranked list, for humans and each model. 
The process starts with a probe, which we gain from the target word $w^r$ and prompt $p^r$ by the verbalization function $\nu$.
Using the probe, we elicit relata $v$ from multiple human participants, or from each model.

We treat the group of humans and each model as a random agent $m$.
During probing experiments, models naturally produce a distribution $D(w^r,p^r;m)$, where each vocabulary item is associated with a probability estimate.
We transform the relata coming from multiple human participants into a single comparable distribution.
We do this by calculating the normalized frequency over relata, after pooling the data coming from different participants. 

From each $D(w^r,p^r;m)$ for either agent $m$, we can create a ranked list $L(w^r,p^r;m)$.
The rank is established by the probability of that word from our distribution over relata $D(w^r,p^r;m)$.
The list returned by models is as long as their vocabulary, so we need to introduce a cutoff $k$, which will be established separately for each metric.
We denote  $L_k(w^r,p^r;m)$ as the resulting response list with the top $k$ words for the agent $m$.
This allows us to treat human responses and model responses in a comparable way. 

\subsection{Soundness and Completeness}

Soundness and completeness are akin to precision and recall.
\textdef{Soundness}, denoted by $\mathcal{S}(r;m)$, is the extent to which words predicted by $m$ are valid relata for relation $r$.

\begin{align}
    \mathcal{S}(w^r;m) &= 
    \frac{1}{\lvert P^r \rvert}\sum_{p^r \in P^r}
    \mathrm{Precision@1}\big(L(w^r,p^r;m),Y^r\big),
    \label{eq:word_sound}\\
    \mathcal{S}(r;m) &= 
    \frac{1}{\lvert W^r \rvert}
    \sum_{w^r \in W^r}
    \mathcal{S}(w^r;m).
\label{eq:soundness}
\end{align}

Note that we first average $\mathrm{Precision@1}$ scores over different prompts for the same target word.
We then average over the target words. 
Previous research~\citep{Ettinger_2020, Ravichander_2020, Hanna_2021} used accuracy, which is mathematically identical to $\mathrm{Precision@1}$, but their numerical values are not comparable to our soundness.
This is because we use relatum sets instead of single gold standard items. 
As any item of our relatum set counts as a hit, soundness values will be generally higher than the accuracy values in the previous work.

\textdef{Completeness} $\mathcal{C}(r;m)$ measures the extent to which $m$ can predict all relata for relation $r$.
\begin{align}
    \mathcal{C}(w^r;m) &= 
    \frac{1}{\lvert P^r \rvert}\sum_{p^r \in P^r}
    \mathrm{Recall@k}\big(L(w^r,p^r;m),Y^r\big),
    \label{eq:word_complete}\\
     \mathcal{C}(r;m) &= 
    \frac{1}{\lvert W^r \rvert}
    \sum_{w^r \in W^r}
    \mathcal{C}(w^r;m).
\label{eq:completeness}  
\end{align}
Completeness averages over the well-known information retrieval metric $\mathrm{Recall@k}$.
Here, we set $k$ to the size of the relatum set or the size of the response list, whichever is smaller.
Soundness and completeness values become identical when the relatum set is singleton.

\subsection{Symmetry}
\label{sec:def_sym}
We measure whether reciprocal elicitation happens for tuple $(w,r,v)$ using metrics proposed by \citet{Mruthyunjaya_2023}~(Equation (\ref{eq:origin_sy})).
Our metrics differ from \citeauthor{Mruthyunjaya_2023}'s in how averaging takes place. 
We average as we do in our calculation of the $\mathcal{S}$ and $\mathcal{C}$ scores.
The summary statistic symmetry $\mathcal{M}_k(r;m)$ is achieved by first obtaining a symmetry score for each tuple, agent $m$, and relation $r$, and then averaging over the tuples. 
\begin{align}
    \mathcal{M}_k(w,r,v;m) &= 
    \frac{1}{\lvert P^r \rvert}\sum_{p^r \in P^r}
    \mu_k(w,v,p^r;m) \label{eq:tuple_symmetry},\\
    \mathcal{M}_k(r;m) &= \frac{1}{\lvert T^r \rvert}
    \sum_{(w,r,v) \in T^r} \mathcal{M}_k(w,r,v;m),
\end{align}
where $r$ is either antonymy or synonymy and $\mathcal{M}_k(w,r,v;m)$ is the symmetry score for tuple $(w,r,v)$.

\subsection{Prototypicality}
We now introduce metrics in order to determine the degree to which prototypicality is observed in the human responses.
Note that unlike in our other metrics, we obtain human experimental performance that establishes the gold standard, rather than some pre-existing lexical data. 

\subsubsection{Response Entropy}
Recall from section~\ref{sec:literature} that prototypicality is defined as the extent to which a particular relata is more exemplary than others, given a relation and a target word.
We can quantify this in the form of the normalized entropy $\mathcal{R}(w^r,p^r)$ of $D(w^r,p^r;h)$, the distribution over relata for target word $w^r$ and prompt $p^r$ produced by humans~($h$)\footnote{Normalization relies on the fact that $\log_2 \lvert D(w^r,p^r;h)\rvert$ is the maximum entropy of a categorical distribution taking $\lvert D(w^r,p^r;h)\rvert$ categories.}. 
\begin{equation}
\begin{aligned}
\label{eq:pe}
    \mathcal{R}(w^r,p^r) = 
        \begin{cases}
         0, & \text{if } \lvert D(w^r,p^r;h)\rvert=1\\
         - \sum_{v\in D(w^r,p^r;h)} \Pr(v)\frac{\log_2\Pr(v)}{\log_2 \lvert D(w^r,p^r;h)\rvert}. & \text{otherwise}
        
        \end{cases}
\end{aligned}
\end{equation}

As for all entropy-based metrics, lower numbers correspond to a stronger prototypicality effect.
Maximal response entropy corresponds to a situation where all participants reply with the same single word, and nothing else. 
We define $\mathcal{R}(w^r,p^r) = 0$ for this case.
Therefore, $\mathcal{R}(w^r,p^r)$ has a range between zero and one. 

\subsubsection{Prototypicality Score}
We evaluate the prototypicality of the model response by comparing it with the human gold standard, which we will establish experimentally later.
In order to realize this, we need a similarity score that rewards models for satisfying the following requirements: 
1) the prototype of the human response, i.e., the word most frequently elicited, is ranked highest in the model's response, and 
2) additionally, in the model's response there are as many other words elicited from humans as possible, with similar rankings.

For example, consider the hypernymy probe \text{``a wall is a part of {\det} {\V}''}.
The human response is [``building'', ``home'', ``house'', ``room'', ...].
Given this ranked list, any model that returns the prototypical holonym \text{``building''} at the top position fulfills the first requirement.
Concerning the other requirement, [``room'', ``building'', ``home'', ``house''] is preferable to [``room'', ``house'', ``home'', ``building''] because it preserves the precedence of \text{``building''} over \text{``home''} and \text{``house''} in the human response.

The first requirement can be implemented using the indicator function $\mathbb{I}[P]$.
The second requirement can be implemented with the edit similarity $E(a,b)$ between word sequences $a$ and $b$, which is based on the edit distance\footnote{The version of edit distance that we use here operates with insertion, deletion and substitution, with a weight of two for substitution and a weight of one for the other operations.
The maximum value of edit distance is the weight of substitution times the length of the larger of the two sequences compared, which is bounded by $2k$. 
We therefore normalize the score by $2k$ and turn it into a similarity metric by reporting the distance from 1.}. 
This time, $k$ differs for each probe; it is set to the number of words in human response for the probe in evaluation. 

We thus define prototypicality $\mathcal{P}(r;m)$ as a distance metric as follows. 
\begin{align}
   \label{eq:prototypicality_instance}
    \rho(w^r,p^r;m) 
    &=\frac{1}{2}\,\mathbb{I}\big[L_1(w^r,p^r;m) =  L_1(w^r,p^r;h)\big]\\
    &\quad+ \frac{1}{2}\,E\big(L_k(w^r,p^r;m), L_k(w^r,p^r;h)\big), \\
    \mathcal{P}(w^r;m) &=
    \frac{1}{\lvert P^r \rvert}\sum_{p^r \in P^r}
    \rho(w^r,p^r;m),\label{eq:word_proto}\\
    \mathcal{P}(r;m) &=
    \frac{1}{\lvert W^r \rvert}\sum_{w^r \in W^r}
    \mathcal{P}(w^r;m). 
\end{align}

The resulting prototypicality metric ranges between zero and one.
A higher value means that a model's response more closely resembles the human response, with a score of one meaning that it is identical to the human response. 

\subsection{Distinguishability}
If a model distinguishes relation $r$ well from relation $s$, then the ranks of $r$-relata in the response should be overall much lower than the ranks of $s$-relata. 

\begin{figure}[ht!]
    \centering
    \includegraphics[width=0.6\linewidth]{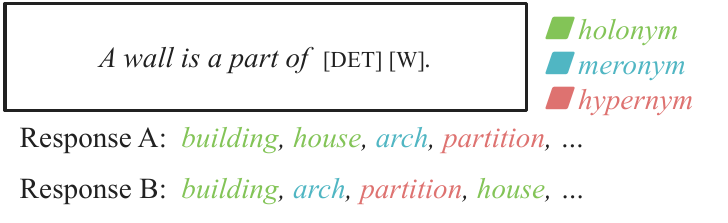}
    \caption{Disinguishability example for a holonymy probe. }
    \label{fig:illustration_disting}
\end{figure}

Consider the example in Figure~\ref{fig:illustration_disting}. 
There are two responses, A and B, to the holonymy probe \text{``A wall is a part of {\det} {\V}''}. 

Holonyms of \text{``wall''} are shown in green.
Note that both responses have correctly placed a holonym in the top rank. 
Despite this, A is intuitively a better response than~B because both holonyms in~A are ranked before all incorrect relata, such as \text{``arch''} (a meronym) and \text{``partition''} (a hypernym).
In contrast, the agent who produced B was less able to distinguish holonymy from meronymy and hypernymy.
Distinguishability was designed to detect the difference between A and B.

\paragraph{Distinguishability Score}
For an ordered pair of semantic relations $(r,s)$, we define $\delta(w^r, p^r, s; m)$ as a mean relative rank of $s$-relata in the response to a probe of relation $r$. In addition, $\delta(r, s; m)$ is defined as the average of $\delta(w^r, p^r, s; m)$ over all prompts and target words. 
The distinguishability of $r$ from $s$, denoted by $\mathcal{D}(r,s;m)$, is the difference between $\delta(r, s; m)$ and $\delta(r, r; m)$ as follows.
 \begin{align}
    \label{eq:confusion}
    \delta(w^r, p^r, s; m) 
    &= \frac{1}{|Y^s|} \sum_{v \in Y^s}{
     \rho\big(v,L_k(w^r,p^r;m)\big),
    }\\
    \delta(r, s;m) &= 
    \frac{1}{\lvert W^r \rvert}\frac{1}{\lvert P^r \rvert}
    \sum_{w^r \in W^r}\sum_{p^r \in P^r}
    \delta(w^r, p^r, s; m),\\
    \mathcal{D}(r,s;m) 
    & = \max \Big(\delta(r, s;m) - {\delta(r,r;m)}, 0\Big),
 \end{align}
where $\rho(a,b)$ is the normalized relative rank of a word $a$ in a list $b$.
Normalization of $\rho(a,b)$ by $k$ (here set to the size of the relatum set of target word $w$) results in a range  $[0,1]$.

Higher $\mathcal{D}$ scores indicate better distinction.
Note that this metric can incur negative values, namely if the highest-ranked correct relatum is ranked after an incorrect relatum. 
In that case, the model has committed an error so grave that we are no longer interested in the rest of the response. 
We therefore assign zero to all cases of negative difference.

Note that our earlier process ensured that all relatum sets are mutually exclusive.
If the intersection between the relatum sets of two relations $r$ and $s$ was not empty, any intersection item would wrongly contribute to both $\delta(r, s;m)$ and $\delta(r,r;m)$.
This leads to a deflation of $\delta(r,s;m)$, meaning that the theoretically highest distinguishability cannot be reached even if an agent were able to perfectly separate $r$-relata from $s$-relata.
The higher the intersection item is ranked, the stronger the negative effect becomes. 

\paragraph{Area under the Distinguishability Curve (AuDC)}
We define the area under the distinguishability curve as a summary statistic for distinguishability.
The distinguishability curve is created as follows. 

\begin{align}
    \eta(p;m) & = \sum_{
    (r,s) \in R \times R\setminus{\{r\}}
    }{
    {\mathbb{I}\Big[\mathcal{D}(r,s;m) > p \Big]}
    }
\end{align}

$\eta(p;m)$ is the number of relation pairs in $R\times R\setminus{\{r\}}$ whose $\mathcal{D}$ score is greater than a threshold $p$.
$p$ can be read as the point at which we are satisfied that agent $m$ successfully distinguishes two relations, with a higher $\eta(p;m)$ requiring better distinguishability.
The distinguishability curve then visualizes the relationship between $p$ and $\eta(p;m)$.

When $p = 0$, all relation pairs with a positive $\mathcal{D}$ score contribute to the $\eta(0;m)$, making it maximal.
As $p$ increases, fewer relation pairs contribute, resulting in a monotonic decrease in $\eta(p;m)$.
At $p = 1$, no relation pairs remain and the curve converges to zero.
Note the similarities to the precision-recall curve in information retrieval, which is also established by varying a threshold. 

The area under the distinguishability curve is obtained as follows.
\begin{align}
    \label{eq:audc}
     \mathrm{AuDC}(m) &= \int^1_0  \eta(p;m) dp
\end{align}
AuDC ranges from zero to the number of all relation pairs, which is 30 in our case.
In contrast to $\eta(p;m)$, which reflects the number of distinguishable relation pairs given a specific $p$, it reflects how many pairs an agent can distinguish on average, with higher numbers meaning higher distinguishability ability. 

\section{Human Experiment}
We now move to the experiments, starting with the human probing experiment. 

\label{sec:human_exp}
\subsection{Elicitation of Human Responses}
In order to collect responses to probes from human participants, we use the Amazon Mechanical Turk~(MTurk) crowdsourcing platform.
Participants were restricted to those 
1) who have the MTurk Master qualification and currently live in either the United States, the United Kingdom, Australia, or Canada, and
2) additionally whose answers are approved more than 500 times at an approval rate above 95\%.
In total, 48 qualified participants were recruited.

We split the 10,507 probes from Table~\ref{tab:relation_statistics} into 276 subsets of 38 probes on average, making sure that no subset contained more than one probe with the same relation and the same target word.
The time limit for responding to each probe was three minutes.
Four participants were assigned to each subset. 
Participants answered 22 subsets on average.

We asked participants to type up to five relata for each probe. 
We instructed them that they should use nouns, but no multi-word expressions. 
We further told participants that the relata could start with either a consonant or a vowel.
In addition to the real probes, we used three additional bogus probes~(such as ``The earth rotates around the {\V}'') per subset, and rejected subsets where the bogus item was not answered correctly (in this case, only ``sun'' was accepted).
All participants correctly answered all bogus probes, so we were able to accept all responses.

The humans responded with 88,896 word tokens (7,077 word types) in total, of which 10,699 word tokens~(12.0\%) are OOR (3,599 OOR types, 51.0\%).
396 response lists consist solely of OOR words~(3.8\%).
On average, we find the first non-OOR words at rank 1.5 in a response list.

On the basis of all responses including OOR, we calculate soundness, completeness, symmetry for antonymy and synonymy, prototypicality, and distinguishability.
Our metrics will penalize agents for responding with an OOR word in each case. 

\subsection{Response Entropy Analysis}
\label{sec:human_pe}
To remind the reader, our gold standard for prototypicality, unlike that for the other metrics, is an outcome of the human experiments, so needs to be calculated before model evaluation can take place. 

We first give some examples of the kinds of prototypes the participants produced. 
For the target word \text{``wall''} under holonymy,  \text{``building''} is the most prototypical item in the response, and for \text{``cloud''}, it is  \text{``sky''}.
These tendencies hold irrespective of which prompt was used.
These two pairs reflect the top-down and bottom-up accounts of holonymy prototype theories (cf. Section~\ref{sec:literature}).

We then look at hypernymy. 
For the target word \text{``orange''}, there are two strong prototypes, namely \text{``fruit''} and \text{``color''}. 
Depending on the probe, they are either tied, or \text{``fruit''} is the most prototypical item, with \text{``color''} being the second.
This aligns with data by \citet{Battig_1969} and \citet{Exbattig}, where more than 80\% of subjects named both \text{``fruit''} and \text{``color''} as hypernyms of \text{``orange''}.

We now address the question whether all relations show a prototypicality effect.
We first consider the responses with the strongest prototypicality, namely, the zero response entropy, where the same single word was the only response of all participants. 

\begin{table}[htp!]
\caption{
Responses with zero response entropy.}
\label{tab:pe_exception}
\begin{tabular}{l|rl|rr}
\toprule
Relation & \multicolumn{2}{c|}{Ratio of responses with $\mathcal{R}$=0} & Length\\
\midrule
Antonymy~(ANT)  & 6.11\% &(50/819)   & 4.28                          \\
Synonymy~(SYN)  & 2.44\% &(36/1477)  & 5.25                          \\
Holonymy~(HOL)  & 1.31\% &(17/1302)  & 5.93                          \\
Hypernymy~(HYP) & 1.14\% &(55/4809)  & 5.59                          \\
Hyponymy~(HPO)  & 0.16\% &(2/1236)   & 7.46                          \\
Meronymy~(MER)  & 0.12\% &(1/864)    & 7.62                          \\
\bottomrule
\end{tabular}
\end{table}
Table~\ref{tab:pe_exception} lists the ratio of such responses, along with the average number of words in responses.
According to this metric, antonymy shows by far the strongest prototypicality at 6.11\% of all responses, followed by synonymy at 2.44\%.
We can observe that meronymy and hyponymy rarely show responses with the strongest prototypicality (only once for meronymy and twice for hyponymy).
This might be related to the fact that these two relations happen to also have more relata in the human responses (more than seven words on average) than other relations, which have an average of around five. 

\begin{figure}[htp!]
    \centering
    \includegraphics[width=0.9\linewidth]{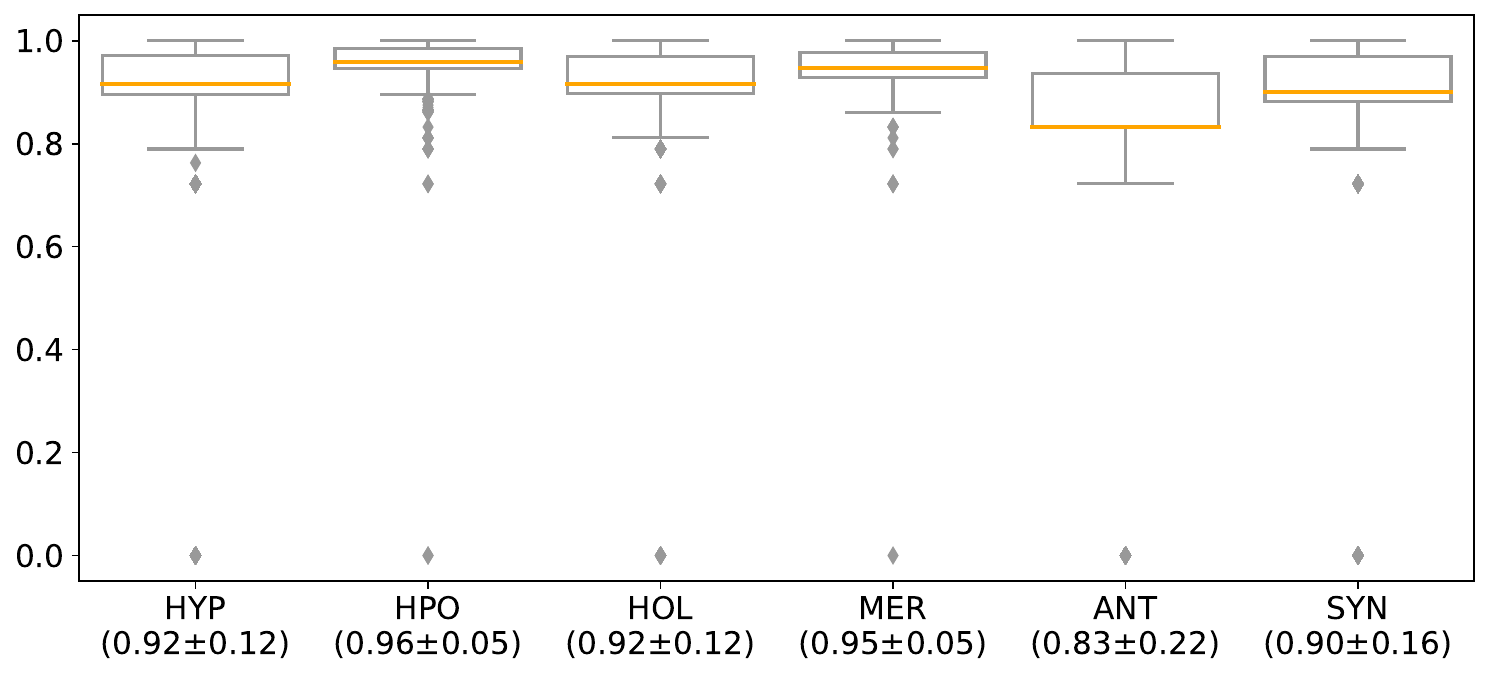}
    \caption{Response Entropy.
    A lower score shows a stronger prototypicality.
    Boxes enclose second and third quantiles, with the mean shown as orange lines. 
    }
    \label{fig:peffect}
\end{figure}

Figure~\ref{fig:peffect} shows the distributions of response entropies $\mathcal{R}$ across relations\footnote{Distributions are significantly different from each other for all relation pairs except for holonymy-hypernymy and holonymy-synonymy relation pairs, as established by Mann-Whitney U tests with $\alpha$=0.05.}. 
We can see that the strongest prototypicality effect by far is again observed for antonymy, at a mean of 0.83. 
Synonymy shows the second strongest prototypicality with a mean of 0.90, with hypernymy and holonymy somewhat less prototypical. 
Hyponymy and meronymy are again at the other extreme, with $\mathcal{R}$ means above 0.95, close to the maximum of 1, and standard deviations lower than those of other relations.  
This means that most distributions for these two relations are close to uniform.
Combining these observations, we conclude that antonymy shows a strong prototypicality effect, but there is hardly any prototypicality effect for hyponymy and meronymy. 
For our planned prototypicality evaluation of the model, we will not use hyponymy and meronymy relations, but only hypernymy, holonymy, antonymy, and synonymy.

If a human response shows a $\mathcal{R}$ of 1, the probe is unable to elicit any prototypical response.
We also remove probes whose responses show a $\mathcal{R}$ of 1.0 from these four relations in the evaluation of prototypicality. 
We find 399 such cases for hypernymy, 197 for hyponymy, 93 for synonymy, 79 for holonymy, 65 for meronymy, and 15 for antonymy.
In addition, words that are out of the model's vocabulary will introduce an artificial deflation in evaluation.
We therefore further discard responses that include any word that is not in the intersection vocabulary of the models tested.
This results in a total of 4,282 responses which we can use in our prototypicality experiments: 2,375 for hypernymy, 798 for holonymy, 447 for antonymy, and 662 for synonymy.

\section{Model Experiment}
\label{sec:model_exp}
For soundness, completeness, and distinguishability, we use all 10,507 probes from Table~\ref{tab:relation_statistics}.
For model prototypicality, we use 4,282 probes, as described above.
The number is higher than 10,507 because of the trick probes we added (cf. Section~\ref{sec:def_sym}). 
For symmetry, we use 1, 5, and 10 for the value of $k$.

\subsection{Target Models}
We use BERT as one of the MLMs because it is widely used in previous work (we use the cased version).
In addition to BERT, we also chose RoBERTa~\citep{roberta}, as this allows us to quantify the effect of training objectives and architectures.
For the CLMs, we chose Llama-3~\citep{LLaMa3}~(from here on, referred to as Llama).
Its monolingual nature ensures the comparability with the two MLMs\footnote{\newcontents{When obtaining the next-token distribution from Llama models, we use the default temperature of 0.6.
However, since we only consider the top-$k$ tokens with the highest probabilities (rather than sampling from the distribution), any positive temperature yields the same evaluation results}.}.

In order to study how model size impacts the learning of semantic relations, we also use models of two different sizes within each model family.
For BERT and RoBERTa, the two variants we use are of similar sizes, but for Llama, the range of sizes we experiment with is much larger, namely one order of magnitude. 

Table~\ref{tab:model_info} lists the statistics of the target models we use. 
\begin{table}[htpb]
    \caption{Statistics of our target models.}
    \centering 
    \begin{tabular}{llrrr}
        \toprule
       Abbr. & Model & \#Parameters 
 
       & \makecell[r]{Vocabulary\\ Size} 
       
       &  \makecell[r]{Pretraining Corpus Size \\(in token numbers)}\\ 
       \midrule
        B1      & BERT-base     &  110M    & 28,996	        &  3.3B \\
        B2      & BERT-large    &  340M    & 28,996	        &  3.3B \\
        R1      &RoBERTa-base   &  125M    & 50,265         & 30B \\
        R2      &RoBERTa-large  &  355M    & 50,265         & 30B \\
        L1      &Llama-8B       &  8B    &  128,256         & 70T \\
        L2      &Llama-70B       &  70B    & 128,256         & 70T \\
 
        \bottomrule
    \end{tabular}
    \label{tab:model_info}
\end{table}

\subsection{Dealing with Determiner Bias}
\label{sec:determiner issue}
Some probes require indefinite determiners before the nouns that are to be predicted.
Keeping determiners fixed in these probes would introduce a bias towards words with an initial vowel or consonant.
For each probe that requires indefinite determiners, we, therefore, probe the models twice, once with ``\text{an}'' inserted into the probe and the other time with ``\text{a}'' inserted into the probe.
From the responses, we create a new distribution over the vocabulary, which is the weighted sum of the two distributions yielded by the two probes.
The weights correspond to the relative frequencies of ``an'' and ``a'' in the Corpus of Contemporary American English~\citep{coca}.
This synthetic distribution can then be treated as the tested model's prediction of the probe.

\subsection{Statistical Test}
Throughout this study, statistical differences in soundness, and symmetry metrics have been tested using McNemar's test with $\alpha=0.05$, as these metrics are binary for each target word or each tuple. 
For completeness and prototypicality, the Wilcoxon signed rank test is used, with $\alpha$=0.05. 
For distinguishability, no known test exists so we do not test for significance.

\section{Results and Analyses}
\label{sec:results}
We present results and analyses based on them.
Before presenting the results of each metric, we examine the general characteristics of responses from each agent by evaluating OOR words. 

\subsection{OOR Words in Responses}
To ensure a fair comparison between humans and models, we identify the OOR words among the top-$k$ words in each model’s response to a given probe, where $k$ corresponds to the number of words produced by humans for the same probe.
Compared to humans, models more frequently generate responses in which all words are OOR. 
Among all models, BERT-large (B2) has the lowest rate of all-OOR responses at 22.3\%, which is much higher than the human rate of 3.8\%.
\newcontents{
In some cases, models do produce OOR words that also appear in human responses. 
For example, humans and all models return ``university'' as a holonym of ``professor'' across six out of seven prompts.
We count the number of OOR words shared between humans and each model for every probe, and then average these counts across all probes.
The results show that the shared OOR words are few, ranging from $0.31$~(Llama-8B) to $0.51$~(RoBERTa-large).
In short, it is rare for models and humans to share the same OOR words.
}

\subsection{Soundness}
\begin{figure}[ht!]
    \centering
    \includegraphics[width=\linewidth]{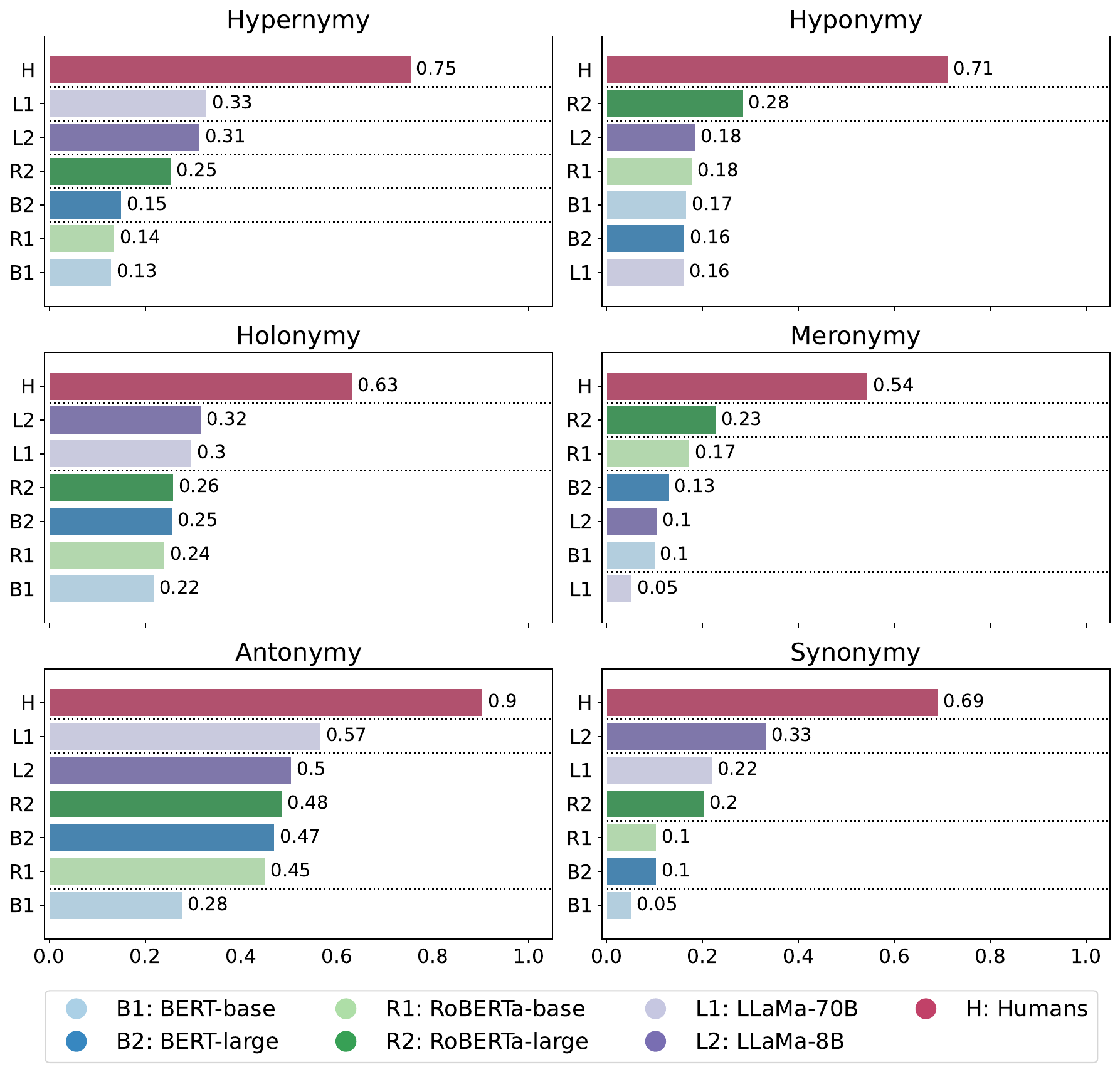}
    \caption{Results of soundness.}
    \label{fig:soundness}
\end{figure}

Figure~\ref{fig:soundness} shows the result of the soundness evaluation for the models, along with the human ceiling. 
In the graphs, we show significance using dotted lines.
A dotted line indicates that the test established statistically significant differences between all models above and below the line\footnote{Note that this is not equivalent to saying that all models between neighbouring dotted lines are statistically indistinguishable. This may or may not be the case for any pair; the notation we use is a simplification in that it cannot express this aspect.}.

\newcontents{Humans (H) show high performance on antonymy ($\mathcal{S}=0.90$), with lower performance for the other relations ($0.54 <\mathcal{S}< 0.75$, with an average of 0.66).}
The performance of all models remains far below that of humans: for most relations, even the best model score is less than half the human score.

\newcontents{All models perform relatively well for antonymy, where the best model Llama-8B~(L1) achieves $\mathcal{S} = 0.57$, whereas for other relations, best values typically lie around $\mathcal{S}=0.30$.}

\newcontents{
When comparing CLMs and MLMs, we can see that CLMs (L1 and L2) outperform MLMs (B1, B2, R1, and R2) for most relations. 
For hypernymy, holonymy, antonymy, and synonymy, either Llama-8B or Llama-70B outperforms all the MLMs.
MLMs have an advantage in hyponymy and meronymy. 
RoBERTa-large (R2) leads all other models with a score of around $\mathcal{S}=0.25$, which is significantly better than that of all other models.}

\subsection{Completeness}
\begin{figure}[ht!]
    \centering
    \includegraphics[width=\linewidth]{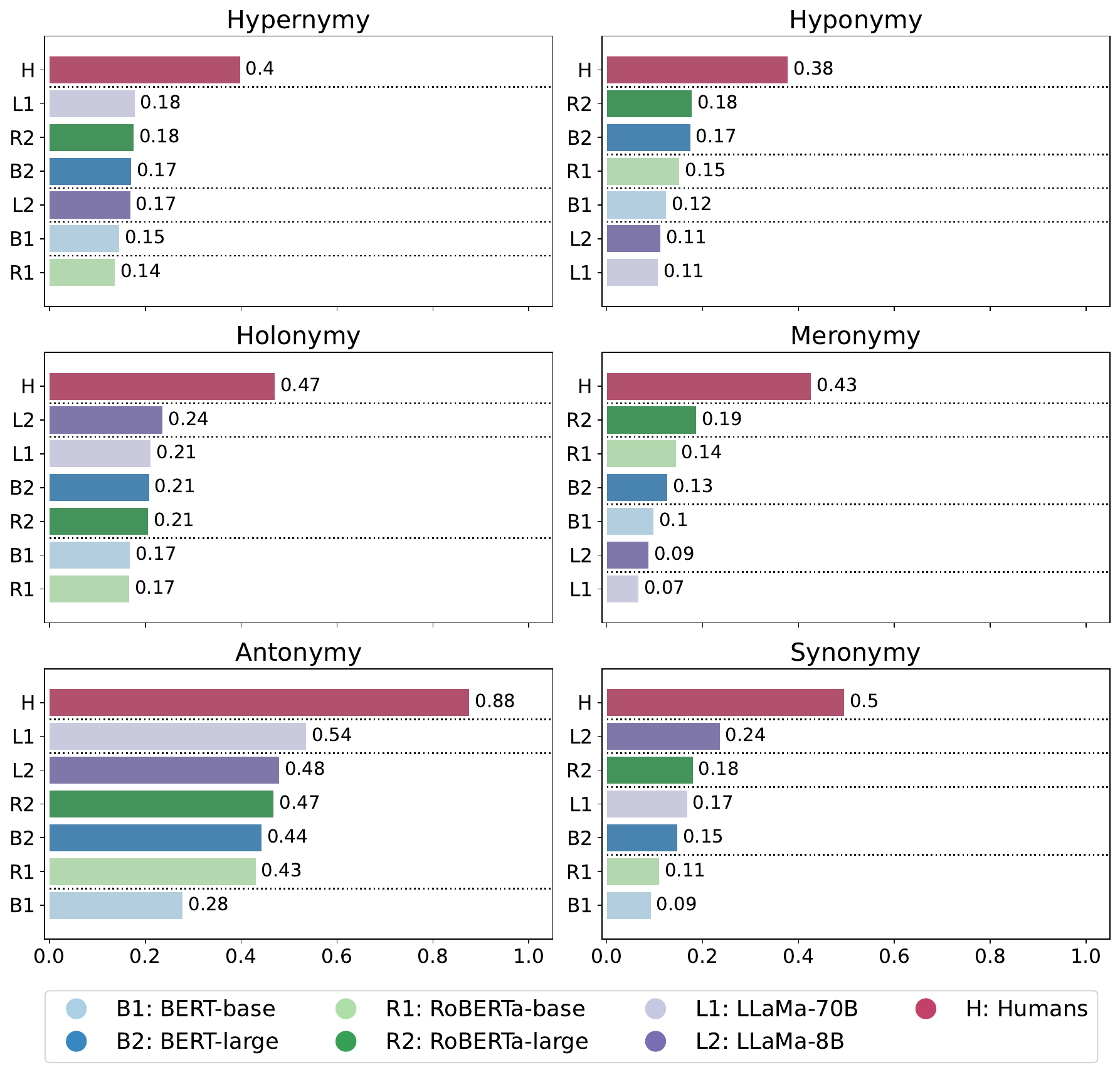}
    \caption{Results of completeness. }
    \label{fig:completeness}
\end{figure}

Figure~\ref{fig:completeness} shows the results of completeness. 
Again, we observe a large gap between models and humans, as was the case for soundness. 
For relations except for antonymy, the human $\mathcal{C}$ scores range from 0.38 to 0.50 and are therefore overall much lower than the human $\mathcal{S}$ scores, where even the lowest score was above 0.50.

\newcontents{The models' $\mathcal{C}$ scores remain below 0.25 for the five relations except for antonymy.
All other trends are similar to those for soundness: 
antonymy stands out again as a relation with high completeness, for both humans and models. 
CLMs' priority can be observed for hypernymy, holonymy, and antonymy but not for hyponymy and meronymy.
For these two relations, the normally observed priority of CLMs is often even reversed: all MLMs outperform CLMs for hyponymy and three out of four MLMs do so for meronymy. 
For synonymy, RoBERTa-large~(R2) shows significant advantage than Llama-8B~(L1) in $\mathcal{C}$ scores, whereas their difference is not significant in $\mathcal{S}$ scores.}

We conclude from the results for the OOR rate, soundness, and completeness that the models only acquire a limited ability to perform relata prediction, which is far below the human ceiling.

\subsection{Symmetry}
\begin{figure}[ht!]
    \centering
    \includegraphics[width=\linewidth]{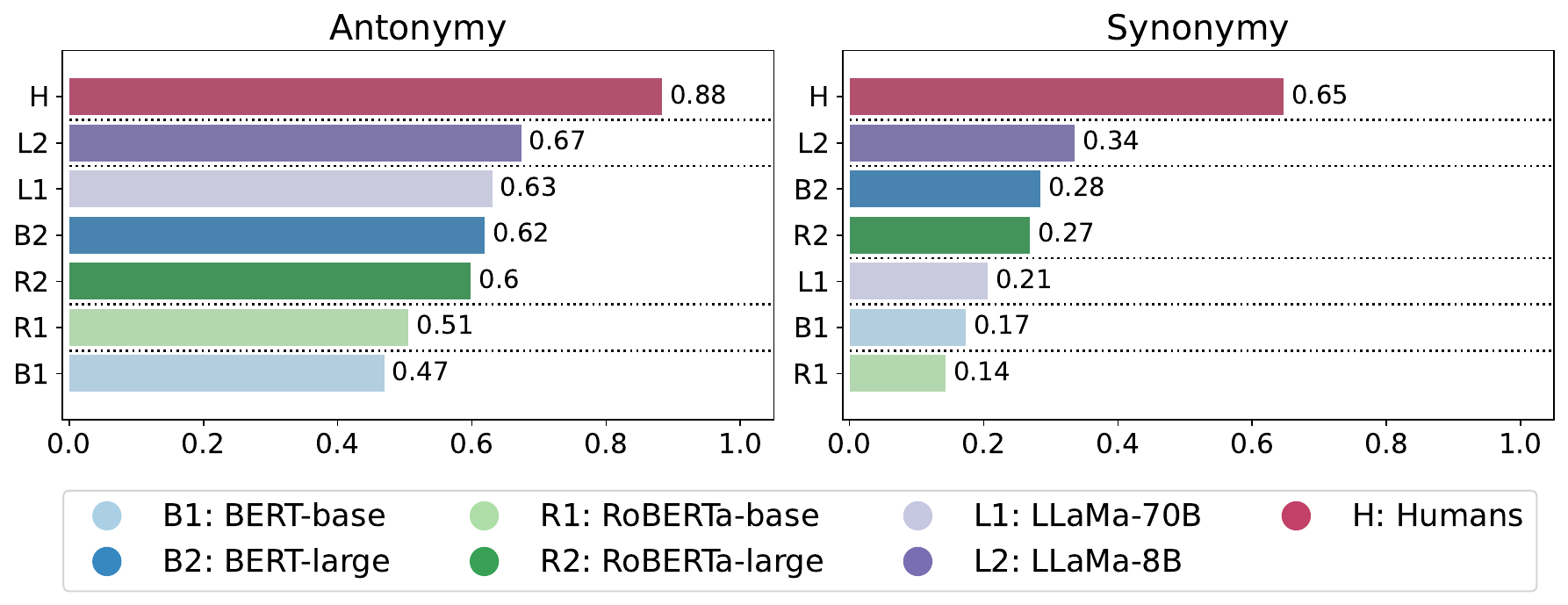}
    \caption{Symmetry.}\label{fig:symmetry}
\end{figure}

Figure~\ref{fig:symmetry} shows the evaluation results for the $\mathcal{M}$ scores with $k=5$\footnote{Results for $k$=1 and $k$=10 are given in Appendix~\ref{app:symmetry_k}. Overall, we observed similar trends for all values of $k$.}.
The humans achieve $\mathcal{M}$=0.88 for antonymy and $\mathcal{M}$=0.65 for synonymy. 
The models similarly perform better for antonymy than they do for synonymy. 
\newcontents{For both relations, the best-perfoming model remains Llama-70B~(L2; $\mathcal{M}=0.67$ for antonymy and $\mathcal{M}=0.34$ for synonymy).
While CLMs generally keep the leading position for antonymy symmetry, MLMs are not generally inferior to CLMs for synonymy symmetry.
BERT-large (B2; $\mathcal{M}=0.28$) and RoBERTa-large (R2; $\mathcal{M}=0.27$) achieve significantly better results than Llama-8B~(L1, $\mathcal{M}=0.21$).
}

In conclusion, all models tested recognize the symmetry of synonymy only to a limited extent. 
In contrast, the best model recognizes the symmetry of antonymy to a relatively high extent, approaching human performance.

\subsection{Prototypicality}
\begin{figure}[ht!]
    \centering
    \includegraphics[width=\linewidth]{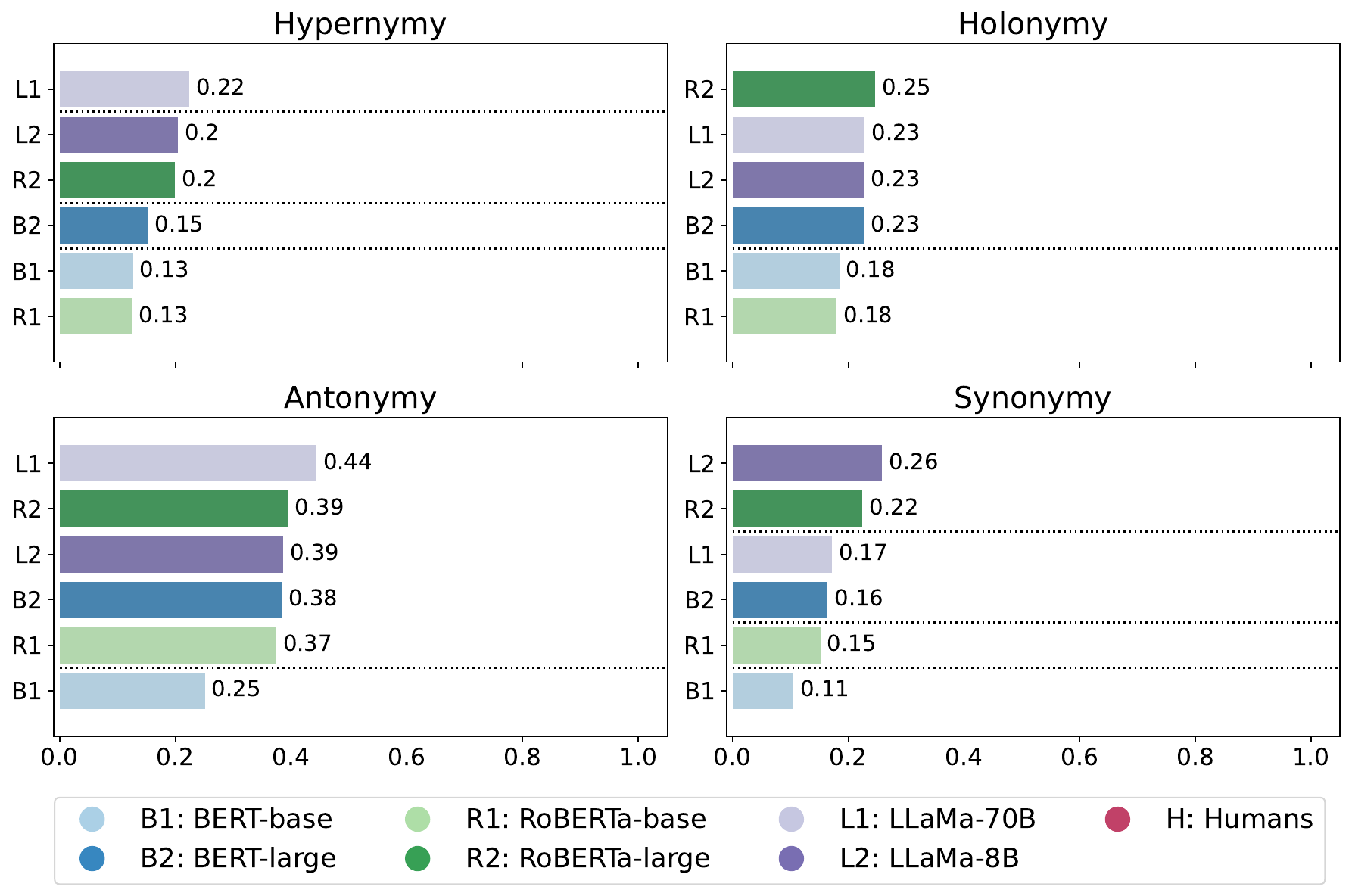}
    \caption{Results for prototypicality. }
    \label{fig:prototypicality}
\end{figure}

Figure~\ref{fig:prototypicality} gives the evaluation results for the $\mathcal{P}$ scores.
Remember that prototypicality is not reported for hyponymy and meronymy, as we have established in Section~\ref{sec:method} that the human responses showed no prototypicality effects for these relations.
\newcontents{For hypernymy, holonymy, and synonymy, all models achieve a $\mathcal{P}$ score below 0.3.
The best-performing models for each relation are as follows.
For synonymy, they are Llama-70B (L2; $\mathcal{P}=0.26$) and RoBERTa-large (R2; $\mathcal{P}=0.22$);
for holonymy, RoBERTa-large (R2; $\mathcal{P}=0.25$), two Llama variants, and BERT-large (L1, L2, B2; all $\mathcal{P}=0.23$);
for hypernymy, Llama-8B (L1; $\mathcal{P}=0.22$).}

\newcontents{Antonymy again is the positive outlier relation, with best results ranging around $\mathcal{P}=0.40$, a value achieved by all models except for BERT-base~(B1).}

\newcontents{In summary, these results indicate that models capture only a limited degree of prototypicality for hypernymy, holonymy, and synonymy, but perform relatively well on antonymy.}

\subsection{Distinguishability}
\begin{figure}[ht!]
    \centering
    \includegraphics[width=0.9\linewidth]{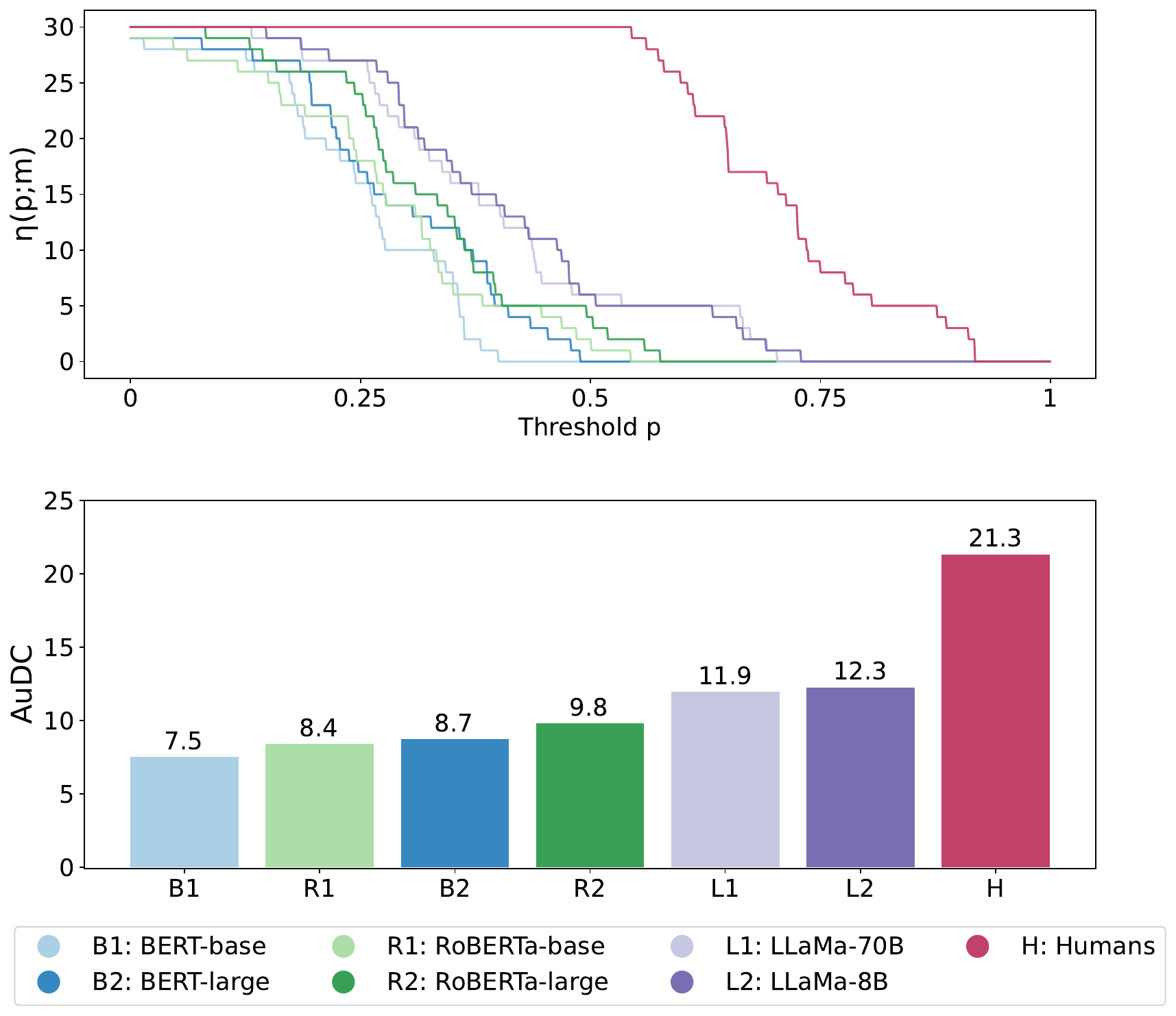}
    \caption{Distinguishability curves~(above) and AuDC~(below).}
    \label{fig:distinguishability_curve}
\end{figure}

Figure~\ref{fig:distinguishability_curve} shows the distinguishability curves for humans and models, with corresponding AuDC values in the lower part of the figure.
We can see in the human results that as $p$ increases, the theoretical maximum of $\eta(p;m)$=30 is kept up until after $p=0.5$.
The curve then gradually descends and reaches zero around $p=0.9$.
\newcontents{The curves for the models show a faster descent than that for humans; zero $\eta(p;m)$ is reached around $p=0.65$ for MLMs and around $p=0.75$ for CLMs.
This means that, although CLMs outperform MLMs, the upper bound for all models remains far below that for humans, suggesting a substantial difference between models and humans in their ability to distinguish relations.}

\begin{figure}[htp!]
    \centering
    \includegraphics[width=\linewidth]{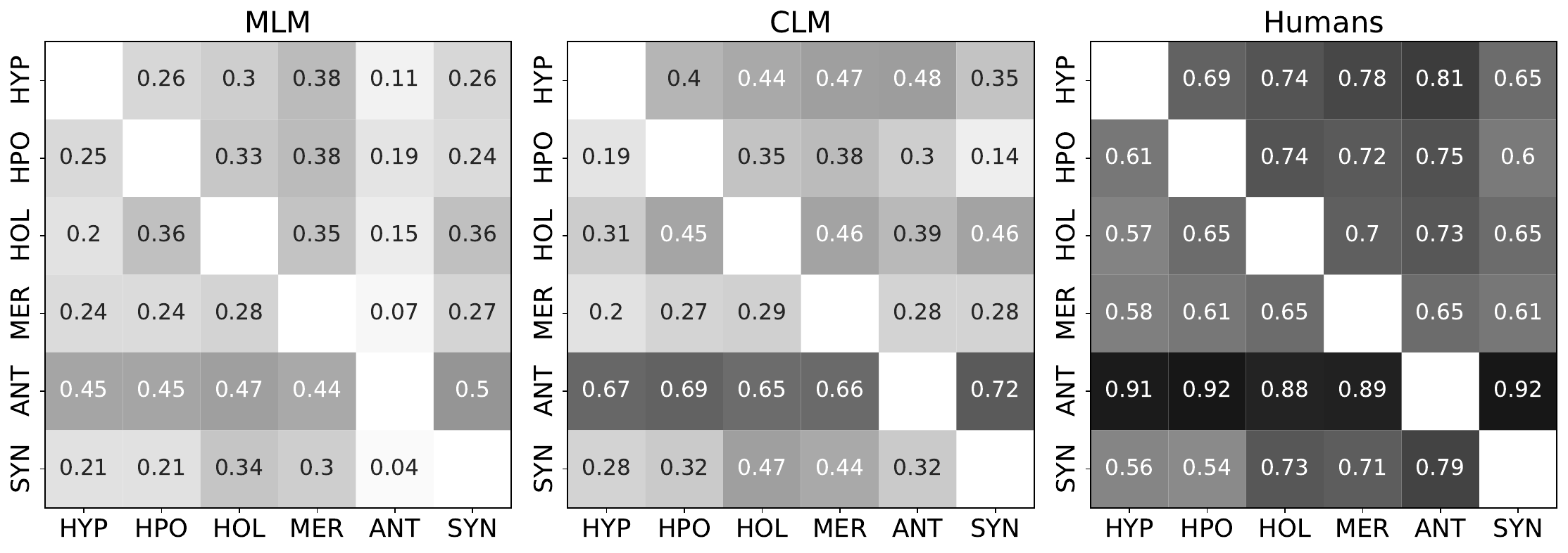}
    \caption{Distinguishability matrices of MLM, CLM, and humans.}
    \label{fig:distinguishability}
\end{figure}
Now let us move on to the lower part of Figure~\ref{fig:distinguishability_curve}, which shows the AuDC values.
The AuDC tells us how many relation pairs out of 30 are distinguished by an agent, on average. 
\newcontents{For humans, the $\mathrm{AuDC}$ value is 21.3, whereas models are only able to distinguish 7.5~(B1) to 12.3~(L2) pairs.
It means that models, at best, would distinguish roughly a third of all relation pairs.
We can therefore safely say that the distinguishability of all models we tested is unsatisfactory. }

Plotting all pairwise $\mathcal{D}$ scores for an agent, we can create the agent's distinguishability matrix. 
This serves to examine more deeply which relations are mistaken for which other ones. 
Figure~\ref{fig:distinguishability} presents distinguishability matrices for entire model families, calculated as averages over all model variations of MLM (left) and CLM (middle), in comparison to the human ceiling (right)\footnote{The complete list of distinguishability matrices for all model variants is provided in Appendix~\ref{app:confumats}.}. 
The relation given at the row position is the prompted relation, and the relation given at the column position is the relation the model responded with.
The depth of grey shade expresses the degree of distinguishability, with lighter cells indicating that the row relation is more often confused with the column relation. 

Humans, despite their generally good ability to distinguish semantic relations, tend to confuse hyponymy and hypernymy with synonymy, as can be seen from the lighter cells at the leftmost bottom ((SYN, HYP) and (SYN, HPO)) and at the rightmost top ((HYP, SYN) and (HPO, SYN)).
This agrees with findings by \citeauthor{Chaffin_1984}: humans perceive synonymy as being close to hypernymy and hyponymy (cf. Section~\ref{sec:literature}).
We also observe that humans' $\mathcal{D}$ scores for hypernymy versus both holonymy and meronymy~(lighter cells, (HYP, HOL) and (HYP, MER) in the leftmost middle) are relatively low as well.
This aligns with the theoretical assumption by \citet{Cruse_1986, Winston_1987} and \citet{Joosten_2010} that there are similarities between holonymy and hypernymy.

\newcontents{
What stands out in both the models' and humans' distinguishability matrices are the high scores for antonymy~(visible in the antonymy row). 
When the prompt is associated with antonymy, humans and models are both able to rank antonyms of the target word before other $r$-relata.
}

\newcontents{
Let us now look at the antonymy \textit{column} in the MLM's matrix.
The antonymy column shows the extent to which the model responds with antonyms although the prompt was associated with a non-antonymy relation.
We can now see a contrasting pattern: for each row, the $\mathcal{D}$ scores in the antonymy column are the lowest of all relations. 
Given these results, we can conclude from this that MLMs generally prefer antonyms of target words, irrespective of the relation prompted.
}

\newcontents{
In contrast, CLMs generally show a weaker preference toward antonyms than MLMs. 
For CLMs, it is never the case that the lowest $\mathcal{D}$ scores appear in the antonymy column.
However, for synonymy prompts, CLMs show mild confusion with antonymy~($\mathcal{D} = 0.32$, the second lowest score in synonymy row).
CLMs tend to struggle with a different distinction, namely that between hyponymy and hypernymy on the one hand, and between hyponymy and synonymy on the other. 
This can be seen from the fact that the respective $\mathcal{D}$ scores are below 0.20~(lighter cells, (HYP, HPO) and (SYN, HPO) in the second row).
}

\newcontents{In summary, the only distinction that the models were able to perform well is that between antonymy and other relations.
Concerning the distinctions between non-antonymy relations, all models perform poorly.
If confronted with an antonymy probe, the models are able to correctly produce antonyms.
However, if confronted with a non-antonymy probe, in some cases, the models even make the mistake of incorrectly producing antonyms.
We now show an example of this confusion.}

\begin{table}[htbp]
\centering 
\caption{Top three words in the prediction of models given the probe ``an answer is similar to {\det} {\V}''}. 
Antonyms are shown in boldface.
\label{tab:case_study}
\begin{tabular}{ll|ccc@{}}
\toprule
Abbr. & Agent          & Top 1       & Top 2       & Top 3 \\
\midrule
H 	&	 Humans 	&	 response	&	reply	& solution \\
\midrule
B1 	&	 BERT-base    	&	 \textbf{question}	&	answer	&	statement \\
B2 	&	 BERT-large 	&	 \textbf{question}	&	statement	&	answer \\
R1 	&	 RoBERTa-base 	&	 \textbf{question}	&	yes	&	answer \\
R2 	&	 RoBERTa-large 	&	 \textbf{question}	&	answer	&	argument \\
L1 	&	 Llama-8B 	&	 \textbf{question}	&	in	&	solution \\
L2 	&	 Llama-70B 	&	 [	&	but	&	\textbf{question} \\
\bottomrule
\end{tabular}
\end{table}

Table~\ref{tab:case_study} shows the different agents' top three relata in response to the probe ``an answer is similar to {\det} {\V}''.
The probe is expected to elicit synonyms of ``answer'': ``response'', ``result'', ``solution'', ``reply'' and ``resolution'', according to our relatum set.
Antonyms of ``answer'', such as ``question'', are shown in boldface. 

\newcontents{
Humans correctly produced only synonyms for this probe, but all models return ``question'' as the first non-OOR response.
All models except for Llama-70B~(L2) return ``question'' as their first response. 
This suggests that the relation of ``question'' and ``answer'' is learned firmly by the models, particularly the MLMs, leading to the problems of confusion between synonymy and antonymy. 
}

\newcontents{
Summing up all the observations so far, we can conclude that antonymy is the relation for which the models show the highest scores in all metrics we introduced: not only for distinguishability, but also for soundness, completeness, symmetry, and prototypicality. 
These lead us to conclude that there must be \textdef{antonymy bias}: models learn only antonymy to a relatively good extent.
}

\subsection{Model Size Analysis}
\label{sec:discussion}
\begin{table}[ht!]
    \caption{
    Performance difference between the large and small variant of three model families, across relations.
}
    \label{tab:model_size_diff}
    \centering
    \begin{tabular}{l|c||rrr}
        \toprule
        Metric & Avg. Type & BERT & RoBERTa & Llama \\
        \midrule
         \multirow{2}{*}{\makecell{Soundness~($\Delta \mathcal{S}$)}}
         & Micro & $+.04$ & $+.09$ & $+.01$  \\
         & Macro & $+.05$ & $+.07$ & $+.02$  \\
        \midrule
         \multirow{2}{*}{\makecell{Completeness~($\Delta \mathcal{C}$)}}
         & Micro & $+.05$ & $+.04$ & $+.01$  \\
         & Macro & $+.06$ & $+.04$ & $+.01$  \\
        \midrule
        \multirow{2}{*}{\makecell{Symmetry~($\Delta\mathcal{M}$)} }
         & Micro & $+.13$ & $+.11$ & $+.10$  \\
         & Macro & $+.13$ & $+.11$ & $+.09$  \\
        \midrule
        \multirow{2}{*}{\makecell{Prototypicality~($\Delta \mathcal{P}$)}}
         & Micro & $+.05$ & $+.07$ & $+.00$ \\
         & Macro & $+.05$ & $+.06$ & $+.00$  \\
        \midrule
        Distinguishability~($\Delta$AuDC)
        & n.a. & $+1.22$ & $+1.40$ & $+0.32$  \\
        \bottomrule
    \end{tabular}
\end{table}
According to the scaling law~\citep{Kaplan_2020}, large-size models should outperform their smaller counterparts, so one should see a positive difference when model size increases.
We examine whether this holds for our tasks.
Table~\ref{tab:model_size_diff} presents the differences in performance between the large and small models in each model family, for all metrics we consider. 
Except for AuDC, all metrics are reported as two different averages. 
Macro differences are averaged over relations, whereas micro differences are averaged over individual scores per target word. 

\newcontents{
As all differences in all metrics shown in Table~\ref{tab:model_size_diff} are positive, the large model outperforms the small model in the same family in almost every case~(except for Llama on prototypicality, where the differences are zero).
However, if there is a performance boost, its magnitude cannot be directly inferred from the model size increase alone.
The same increase in size~(330 million, 110M to 340M for BERT and 125M to 355M for RoBERTa) generally affords RoBERTa a higher degree of improvement in soundness, prototypicality, and distinguishability than it affords BERT.
Despite the smaller increase in model size of BERT and RoBERTa, when compared to that of Llama (72 billion), the former two MLMs consistently exhibit a greater performance boost than Llama, where the improvements are overall marginal.
}

\newcontents{
When the performance differences are broken down into individual relations, we find that the relations that benefit from the increase in model size vary across model families\footnote{Results are shown in Appendix~\ref{app:modelsize}).}. 
Antonymy shows the highest improvement among all relations for BERT, ranging from 0.14 to 0.19.
For RoBERTa, the highest improvement is usually observed in hypernymy or synonymy, ranging between 0.08 to 0.12.
In contrast, increasing the size of Llama does not consistently yield better performance.
For hypernymy and antonymy, the smaller Llama-8B generally significantly outperforms the larger Llama-70B across almost all metrics\footnote{Except for antonymy symmetry, the boost in scores is only 0.04.}. 
For holonymy soundness and prototypicality, as well as hyponymy completeness, there is no significant improvement.
}

\newcontents{
To sum up, while large models typically outperform small models, we have seen enough counterexamples to conclude that this regularity does not necessarily or even generally hold.
Therefore, model size is not all that matters for the learning of semantic relations.
}

\subsection{Pretraining Task Analysis}
In factual probing tasks, MLMs have been shown to outperform CLMs~\citep{Petroni_2019, Cao_2022, Mruthyunjaya_2023} whereas currently CLMs are used more wildly in a serious of tasks~\citep{opt,llama,LLaMa3}, as we have discussed in Section~\ref{sec:literature}.
We now verify which pretraining task leads to a better quality of semantic relations knowledge.

We first calculate differences between pairs of the best-performing CLM and MLM per metric. 
The best-performing MLM is always either RoBERTa-large~(R2) or BERT-large~(B2), whereas the best CLM could be either Llama-8B(L1) or Llama-70B(L2); different conditions produce different pairs. 
The results are presented in Table~\ref{tab:CLM_superiority_best}.

 \begin{table}[htbp]
 \caption{Difference between the best models pretrained on different tasks~(the best CLM minus the best MLM). Statistically significant within-metric differences are reported as values; non-significant differences are parenthesized.
 }
 \label{tab:CLM_superiority_best}
 \centering
     \begin{tabular}{l|rrrr}
     \toprule
     Relation 
     & \makecell[r]{Soundness\\ $\Delta \mathcal{S}$}
     & \makecell[r]{Completeness\\ $\Delta \mathcal{C}$} 
     & \makecell[r]{Symmetry\\ $\Delta\mathcal{M}$} 
     & \makecell[r]{Prototypicality\\ $\Delta \mathcal{P}$}\\

     \midrule
     HYP & $+$.07 & (.00) & n.a. & $+$.02  \\
     HPO & $-$.10 & $-$.07 & n.a. & n.a. \\
     HOL & $+$.06 & $+$.03 & n.a. & $-$.02 \\
     MER & $-$.12 & $-$.10 & n.a. & n.a. \\
     ANT & $+$.08 & $+$.07 & $+$.05 & (.03) \\
     SYN & $+$.13 & $+$.06 & $+$.05 & (.03) \\
     \bottomrule
     \end{tabular}
 \end{table}

As there are positive, negative, and non-significant performance differences in Table~\ref{tab:CLM_superiority_best}, it is never the case that the best CLM always outperforms the best MLM.
For AuDC, we find that the best-performing pair is Llama-70B~(L2) and RoBERTa-large~(R2), with a numerical difference of 2.5 in favour of the CLM.

\newcontents{For prototypicality, there is either no significant difference at all, or if there is one, it is small. 
Combining this observation with the fact that even the best performance on prototypicality is below 0.44~(achieved by Llama-8B for antonymy), we can conclude that both types of models do not perform well under this metric.}

For soundness and completeness of hyponymy and meronymy, the best CLM significantly falls behind the best MLM by at least 0.07.
\newcontents{However, for soundness and completeness of hypernymy and holonymy, which are the reverse relations of hyponymy and meronymy, the best CLM shows either comparable or significantly better performance than the best MLM.
In other words, CLMs are more sensitive to relation directionality.
}
This sensitivity aligns with the \textdef{reversal curse} observed in the previous literature~\citep{Berglund_2024}, where CLMs only solve factual prompting tasks in one direction but not in the reverse direction.
We want to point out that in all cases, the CLM model is larger than the MLM model. 
Even the \textit{smallest} size difference observed between the pair of best MLM and best CLM~(RoBERTa-large and Llama-8B) is more than 6 billion parameters.
Our results further show that this sensitivity is so strong that, when the direction is reversed, CLMs' performance drops below that of MLMs, which are tens of times smaller in size.

To gather further evidence, we compare the performance of every CLM against every MLM. 
In other words, for every MLM-CLM pair~(8 pairs in total), we determine if the MLM significantly outperforms the CLM.
We then count the number of pairs where the MLM is significantly better.
Table~\ref{tab:significant pairs} shows the results.

\begin{table}[htbp]
 \caption{Number of MLM-CLM pairs where the MLM significantly outperforms the CLM, out of eight pairs.}
 \label{tab:significant pairs}
 \centering 
     \begin{tabular}{l|rrrr}
     \toprule
     Relation& Soundness & Completeness. & \makecell[r]{Symmetry} & Prototypicality 
     \\
     \midrule
     HYP      & 0 & 2   &  n.a   & 0     \\
     HPO      & 2 & 8   &  n.a & n.a.     \\
     HOL      & 0 & 0   &  n.a & 1      \\
     MER      & 6 & 7   &  n.a & n.a.    \\
     ANT      & 0 & 0   & 0  & 0      \\
     SYN      & 0 & 1   & 2  & 1      \\
     \bottomrule
     \end{tabular}
 \end{table}
 
The significant advantage of MLM can be observed particularly in hyponymy and meronymy, where at least 2 out of 8 MLM-CLM pairs belong to such cases\footnote{
For AuDC, as there is no statistical test, we compare model pairs numerically. 
We do not find any MLM outperforming a CLM.}.  
For hyponymy and meronymy completeness, almost all CLMs remain below MLMs.

Hence, this confirms 
that there is no general superiority of CLMs over MLMs, 
and that CLMs present weakness in learning hyponymy and meronymy, which is hard to even be bridged by increasing model sizes.

\subsection{Word Frequency Analysis}
BERT is known to achieve higher accuracy scores in hypernym prediction tasks when the target word is frequent~\citep{Ravichander_2020}; it is therefore prudent to perform a correlation analysis of results and word frequency.
If the frequencies of target words and relata are partially responsible for the performance of a model, there should be a positive correlation between performance and frequency.

For every model, we use the rank correlation metric Spearman's $\rho$ to compare word-frequency metrics that we derive independently from \newcontents{COCA~\citep{coca}}, against all metrics except AuDC.
We exclude AuDC from this analysis, as AuDC is a metric that is not lexically determined. 
As soundness, completeness, and prototypicality are summary statistics calculated across target words, we first need to recompile individual scores per target word.
These scores can be obtained using Equations~(\ref{eq:word_sound}), (\ref{eq:word_complete}) and (\ref{eq:word_proto}).
We then calculate correlations between the scores of these three metrics with word frequencies.

Soundness and completeness are relations that involve relatum sets.
We therefore also need to consider the frequency of relata, not only of target words\footnote{Prototypicality scores do not use relatum sets as gold standard. 
Thus, we only report correlation with target words for prototypicality.}.
We calculate the correlation between the scores against the average and maximum frequency of relata in each gold relatum set.
We choose the average and maximum because we need to consider the relatum set as a whole, as this is how the sets are used in the calculation of these metrics.

Symmetry requires special treatment because it is defined on tuples involving two words (the target word and the relatum). 
Agents might be unfairly advantaged in recognizing symmetry by word frequency effects in two cases:  
1) if both words in the tuple are common and 
2) if one word in the tuple is far more common than the other.
We use two frequencies:  
average frequency, which can guard against the first case, 
and absolute frequency difference, which can guard against the second. 
We first recalculate the symmetry scores per tuple, using Equations~(\ref{eq:tuple_symmetry}), and then correlate them with the average frequency and the absolute frequency difference.

This allows us to determine correlation for certain metric and relation combinations:
For soundness and completeness, there are coefficients for all six relations.
For symmetry, there are coefficients for the two symmetric relations.
For prototypicality, there are coefficients for hypernymy, holonymy, antonymy, and synonymy. 
Consequently, the total number of coefficients is $6\times (2\times 6 + 1\times 2 + 1\times 4)=60$.
The results are that correlations for all models have medians below 0.30 across all metrics and relations considered\footnote{Details can be found in Appendix~\ref{app:freq}.}.
\citet{Hinkle_2003} regard correlations below 0.30 as negligible.

\newcontents{
As the metrics used for frequency are based on unigram, we conclude that the unigram frequency is unlikely to be the decisive factor in semantic relation learning. 
}

\section{Limitations}
Our work has some methodological limitations.
We adopt prompt-based probing as our core method, but all prompt-based methods suffer from a high dependency on specific prompt design~\citep{Ravichander_2020,Elazar_2021, Cao_2021}. 
We counteract this dependency by using several different prompts for each semantic relation, but we cannot be sure that this is enough. 
\citet{Cao_2022} presents a method for the mitigation of prompt dependency in evaluations, which we implemented and applied to our results, but which resulted in little difference\footnote{All results reported in the paper were therefore given in their original form, i.e., without mitigation.}. 

A possible way to investigate probe dependency more thoroughly is to determine whether the scores from different prompts are heteroscedastic or homoscedastic. 
\textit{Heteroscedasticity} is the property of several samples to have a different variance~\citep{Levene_1974}.
Preliminary results of this analysis are given in Appendix~\ref{app:prompts_hetero}.
We found that for all models, some prompts present heteroscedasticity under certain metrics, but for humans, the evidence supporting heteroscedasticity is insufficient. 

In the general case, prompt dependency remains an unsolved question.
We suspect that some models may use linguistic expressions in certain prompts as shortcuts when solving semantic relation task.
For improving the evaluation methodology presented here, the identification of such shortcut expressions is the next step for the mitigation of prompt dependency, which should secure more stable results.

\newcontents{
We chose to perform our evaluation at the word level, rather than the subword level. 
This decision was made in order to ensure the comparability between CLMs and MLMs.
The number of masks that MLMs need to fill has to be determined in advance.
However, CLMs can predict a sequence of subwords of any length.
The pre-defined numbers of masks may provide extra hints to MLMs, which may give an advantage to them and hence may work against comparability.
}

In our word frequency analysis above, we found little correlation between scores and word frequency, but we only examined the unigram frequency of individual words.
Unigram frequency cannot use any information of syntagmatic or paradigmatic relationships between two words.
Using frequency-based metrics that can exploit such information may lead to a different conclusion.
Particularly, more complex frequency measures that capture syntagmatic or paradigmatic relationships would be desirable.
However, they require a precise definition of such measures and carefully controlled experiments, which are aspects beyond the scope of the present study.

\section{Conclusion}
\label{sec:conclusion}

Current PLMs are commonly used for a wide range of tasks, so it is important to gain a comprehensive understanding of their linguistic abilities.
This study focuses on the semantic relation knowledge of PLMs.
In particular, it explicates the gap in semantic relation knowledge between current PLMs and humans.
Our contributions are as follows. 
\begin{enumerate}
   \item We presented a prompt-based probing evaluation methodology that covers five aspects of semantic relation knowledge, namely soundness, completeness, symmetry, prototypicality, and distinguishability. 
    Two of these metrics are novel, namely those for prototypicality and distinguishability. 
    We also employed established metrics in a new context. 
    \item Using these evaluation methods, we conducted a comprehensive evaluation of the above-mentioned aspects for six nominal semantic relations, five of which were underexplored in probing experiments before. 
    \item We established the first human gold standard for prototypicality.
    For the other aspects, where we constructed gold standards from existing lexicographic data, we established the first human ceiling, which can be used in comparisons with automatic models. 
\end{enumerate}

Our experiment afforded far more conditions and distinctions than previous studies, including a human ceiling and a comparison of CLMs with MLMs. 
As a result, we arrive at a richer characterization of PLMs' capabilities with semantic relations than was possible before. 
Our main result is that PLMs fall short of achieving human-level performance on the extensive semantic relation tasks defined here. 

We experimentally studied the prototypicality effect as displayed by humans for the six relations.
We were able to confirm prototype effects for hypernymy and antonymy that have been experimentally studied in the literature.
We also found a prototype effect for holonymy and synonymy that was not known before.
One of our most important findings for prototypicality was that hyponymy and meronymy showed little prototypicality. 

\newcontents{We also found that a large model size does not always guarantee better performance in learning semantic relations.
The type of PLMs also matters.
MLMs consistently outperformed CLMs for hyponymy and meronymy.
Therefore, the bidirectional context utilized by MLMs is a crucial factor in learning these two semantic relations.}

\newcontents{We observed an antonymy bias: both humans and models performed best for antonymy out of all relations; the results are stable across all metrics. 
MLMs often misrecognize non-antonymy relations as antonymy.
On the other hand, CLMs only misrecognize synonymy as antonymy.
Humans performed well with antonymy in both directions.
Since the antonymy bias appears across all the models evaluated, it may not be attributed to the pretraining tasks or the model sizes.
One possible reason might be the distributional characteristics of the antonymy relation in the corpus, which make antonymy fundamentally different from the other relations.
Antonym pairs commonly co-occur in conjunctive structures such as ``ascent and descent'' and ``day or night''.
The other relations might not present such strong co-occurrence tendencies.
This contrast could explain the antonymy bias we observed in the present study.
Future work includes studying how the co-occurrences of antonyms affect their recognition by models.
Such insights should also lead to better learning methods for the other relations.
}

Our study aims to contribute towards a future where PLMs can better understand language. 
The fact that PLMs struggle to understand several aspects of semantic relations contradicts the superiority of PLMs as observed in many NLP tasks. 
This superiority is commonly attributed to the assumption that PLMs are able to efficiently encode general semantic and linguistic knowledge. 
Such a strong assumption requires solid evaluation methodology to be able to formulate them scientifically and substantiate it. 
Our methodology is able to accomplish this.
Through the comprehensive evaluation, our work puts doubts on this assumption.
Some other explanations for the good performance should be sought. 
In general, it captures effects not seen before. 
We, therefore, consider it a prism through which to see the truth more clearly.

\appendix
\newpage
\begin{appendices}
\section{All Prompts}
\label{app:prompts}

\begin{table}[hb!]
\caption{All prompts used in this research. Presented per relation.}
\label{tab:all_prompts}
    \centering
    \begin{tabular}{l|l}
    \toprule
    Relation & Prompt \\
    \midrule
        \multirow{7}{*}{HYP~(7)}
        &  {\det}  {\W} is a type of  {\det}  {\V} \\
        &  {\det}  {\W} is a kind of  {\det}  {\V} \\
        &  the word  {\W} has a more specific meaning than the word  {\V} \\
        &  {\det}  {\W} is  {\det}  {\V} \\
        &  {\det}  {\W} is a specific case of  {\det}  {\V} \\
        &  {\det}  {\W} is a subordinate type of  {\det}  {\V} \\
        &  the word  {\W} has a more specific sense than the word  {\V} \\
        \midrule
        \multirow{4}{*}{HPO~(4)} 
        &  my favorite  {\W} is  {\det}  {\V} \\
        &  {\det} W, such as  {\det}  {\V} \\
        &  the word  {\W} has a more general meaning than the word  {\V} \\
        &  the word  {\W} has a more general sense than the word  {\V} \\
        \midrule
        \multirow{7}{*}{HOL~(7)}
        &  {\det}  {\W} is a component of  {\det}  {\V} \\
        &  {\det}  {\W} is a part of  {\det}  {\V} \\
        &  {\det}  {\W} is contained in  {\det}  {\V} \\
        &  {\det}  {\W} belongs to constituents of  {\det}  {\V} \\
        &  {\det}  {\W} belongs to parts of  {\det}  {\V} \\
        &  {\det}  {\W} belongs to components of  {\det}  {\V} \\
        &  {\det}  {\W} is a constituent of  {\det}  {\V} \\
        \midrule
        \multirow{6}{*}{MER~(6)} 
        & constituents of  {\det}  {\W} include  {\det}  {\V} \\
        & components of  {\det}  {\W} include  {\det}  {\V} \\
        & parts of  {\det}  {\W} include  {\det}  {\V} \\
        &  {\det}  {\W} consists of  {\det}  {\V} \\
        &  {\det}  {\W} has  {\det}  {\V} \\
        &  {\det}  {\W} contains  {\det}  {\V} \\
        \midrule
        \multirow{9}{*}{ANT~(9)}
        &  it is not likely to be both  {\det}  {\W} and  {\det}  {\V} \\
        &  {\det}  {\W} is the opposite of  {\det}  {\V} \\
        &  the word  {\W} has an opposite sense of the word  {\V} \\
        &  it is impossible to be both  {\det}  {\W} and  {\det}  {\V} \\
        &  the word  {\W} has a meaning that negates the meaning of the word  {\V} \\
        &  it is  {\det}  {\W} so it is not  {\det}  {\V} \\
        &  the word  {\W} has an opposite meaning of the word  {\V} \\
        &  if something is  {\det} W, then it can not also be  {\det}  {\V} \\
        &  the word  {\W} has a sense that negates the sense of the word  {\V} \\
        \midrule
        \multirow{7}{*}{SYN~(7)}
        &  {\det}  {\W} is also known as  {\det}  {\V} \\
        &  {\det}  {\W} is often referred to as  {\det}  {\V} \\
        &  the word  {\W} has a similar meaning as the word  {\V} \\
        &  {\det}  {\W} is similar to  {\det}  {\V} \\
        &  the word  {\W} means nearly the same as the word  {\V} \\
        &  {\det}  {\W} is indistinguishable from  {\det}  {\V} \\
        &  {\det}  {\W} is also called  {\det}  {\V} \\
    \bottomrule
    \end{tabular}
\end{table}

\clearpage
\section{Results of Symmetry}
\label{app:symmetry_k}
\begin{figure}[ht!]
        \centering
        \includegraphics[width=0.9\linewidth]{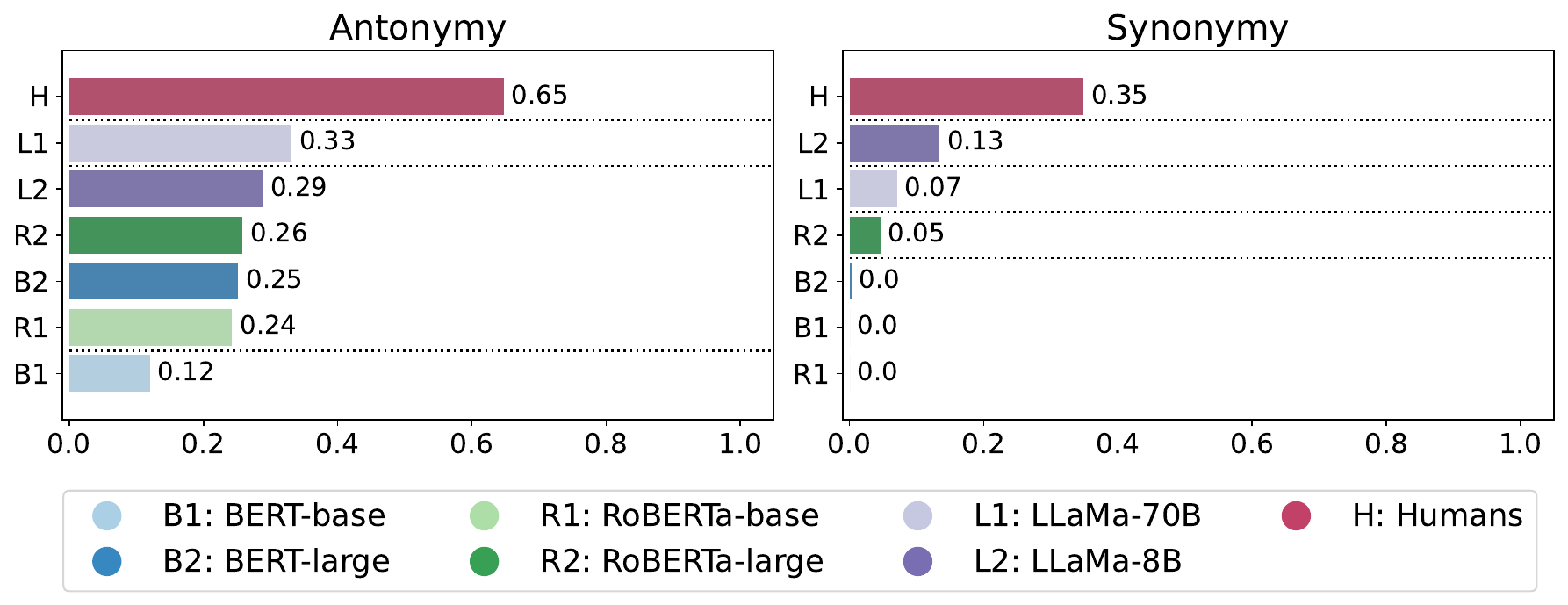}
        \caption{Results for symmetry when $k=1$.}\label{fig:symmetry_1}
\end{figure}
\begin{figure}[ht!]
        \centering
        \includegraphics[width=0.9\linewidth]{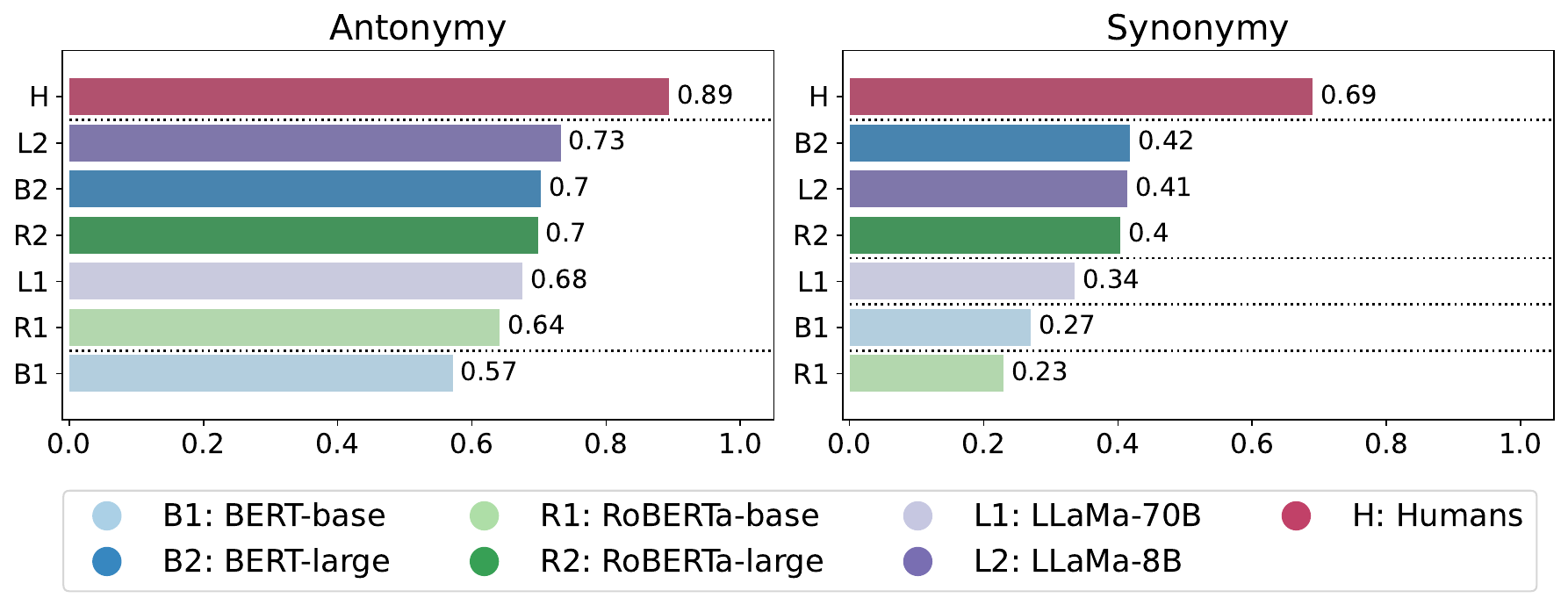}
    \caption{Results for symmetry when $k=10$.}\label{fig:symmetry_10}
\end{figure}

\clearpage
\section{Confusion Matrices}
\label{app:confumats}
\begin{figure}[H]
    \centering
    \includegraphics[width=0.8\linewidth]{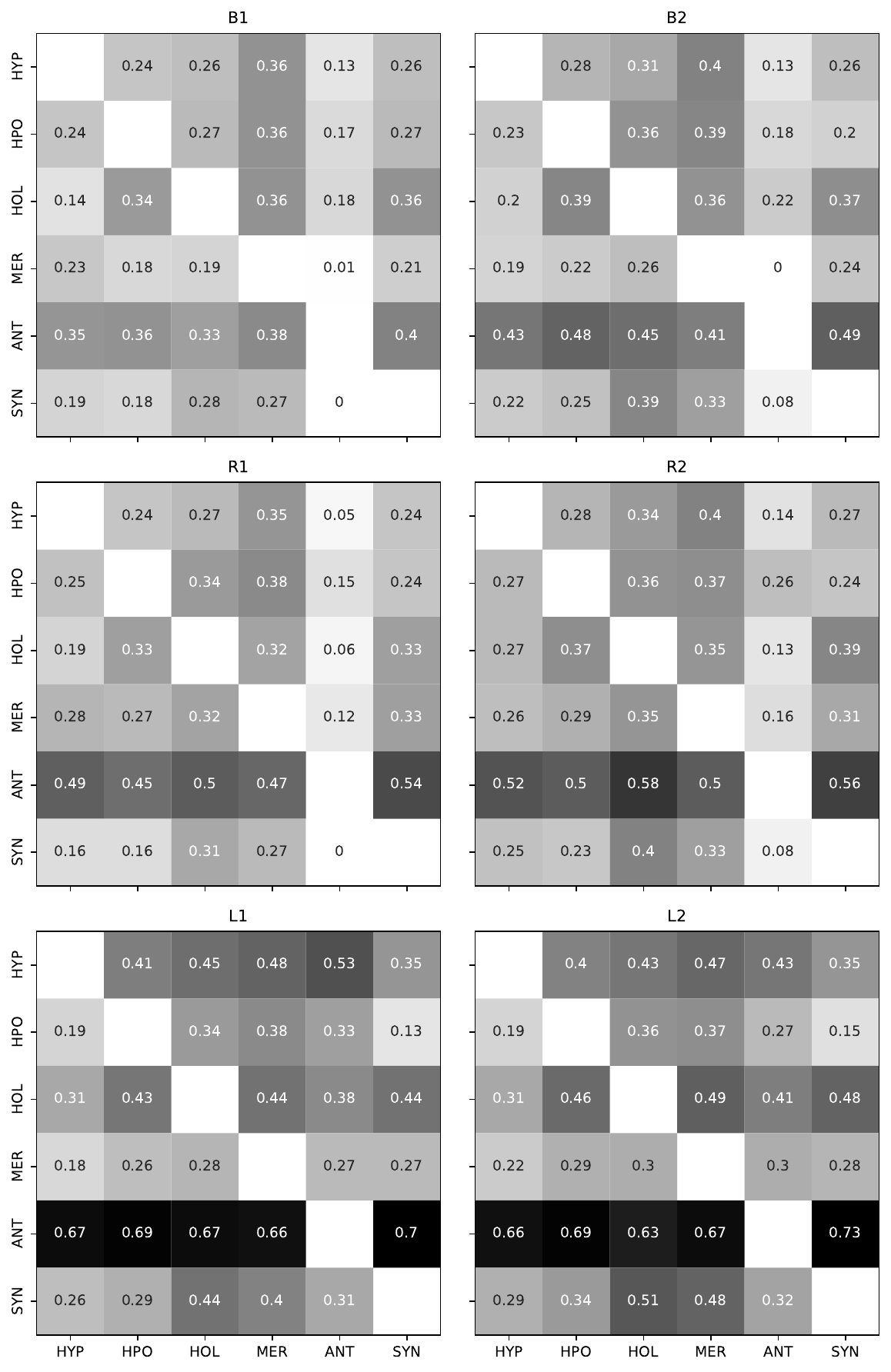}
\caption{Distinguishability matrices of all Models}
\label{fig:allconf}
\end{figure}

\clearpage
\section{Model Size Difference per Relation and Metric}
\label{app:modelsize}
Table~\ref{tab:detailed model size diff} shows the performance difference between the largest and the smallest model for four model families.
\begin{table}[ht!]
    \caption{
    Performance difference between the largest and the smallest model for four model families.
    All differences in metrics are significant unless they appear in a bracket. 
    The maximum within model families and metrics is boldfaced.
}
    \label{tab:detailed model size diff}
    \centering
    \begin{tabular}{l|c||rrrr}
        \toprule
        Metric & Relation & BERT & RoBERTa & Llama \\
        \midrule
         \multirow{6}{*}{Soundness}
         & HYP &  .02  &  \textbf{.12}  &  $-$.01  \\
         & HPO &  (.00)  &  .11  &  .02  \\
         & HOL &  .04  &  (.02)  &  (.02)  \\
         & MER &  .03  &  .05  &  .05  \\
         & ANT &  \textbf{.19} &  .04  &  -.06   \\
         & SYN &  .05   &  .10  &  \textbf{.11}  \\
        \midrule
         \multirow{6}{*}{Completeness}
         & HYP &  .02  &  .04  &   $-$.01  \\
         & HPO &  .05  &  .03  &  (.00)  \\
         & HOL &  .04  &  .04  &  .02  \\
         & MER &  .03  &  .04  &  .02  \\
         & ANT &  \textbf{.16} &  .04  &  $-$.06 \\
         & SYN &  .06  &  \textbf{.07}  &  \textbf{.07} \\
        \midrule
        \multirow{2}{*}{Symmetry}
         & ANT &  \textbf{.15}  &  .09 &  .04 \\
         & SYN &  .11           &  \textbf{.13}  &  \textbf{.13} \\
        \midrule
        \multirow{4}{*}{Prototypicality}
         & HYP &  .03  &  \textbf{.08}  &  $-$.02 \\
         & HOL &  .05  &  .07  &  (.00) \\
         & ANT &  \textbf{.14}  &  (.03) &  $-$.05  \\
         & SYN &  .05  &  \textbf{.08}  &  \textbf{.07} \\
        \bottomrule
    \end{tabular}
\end{table}

\clearpage
\section{Word Frequency Correlation}
\label{app:freq}
\begin{figure}[htpb]
    \centering
    \includegraphics[width=0.9\linewidth]{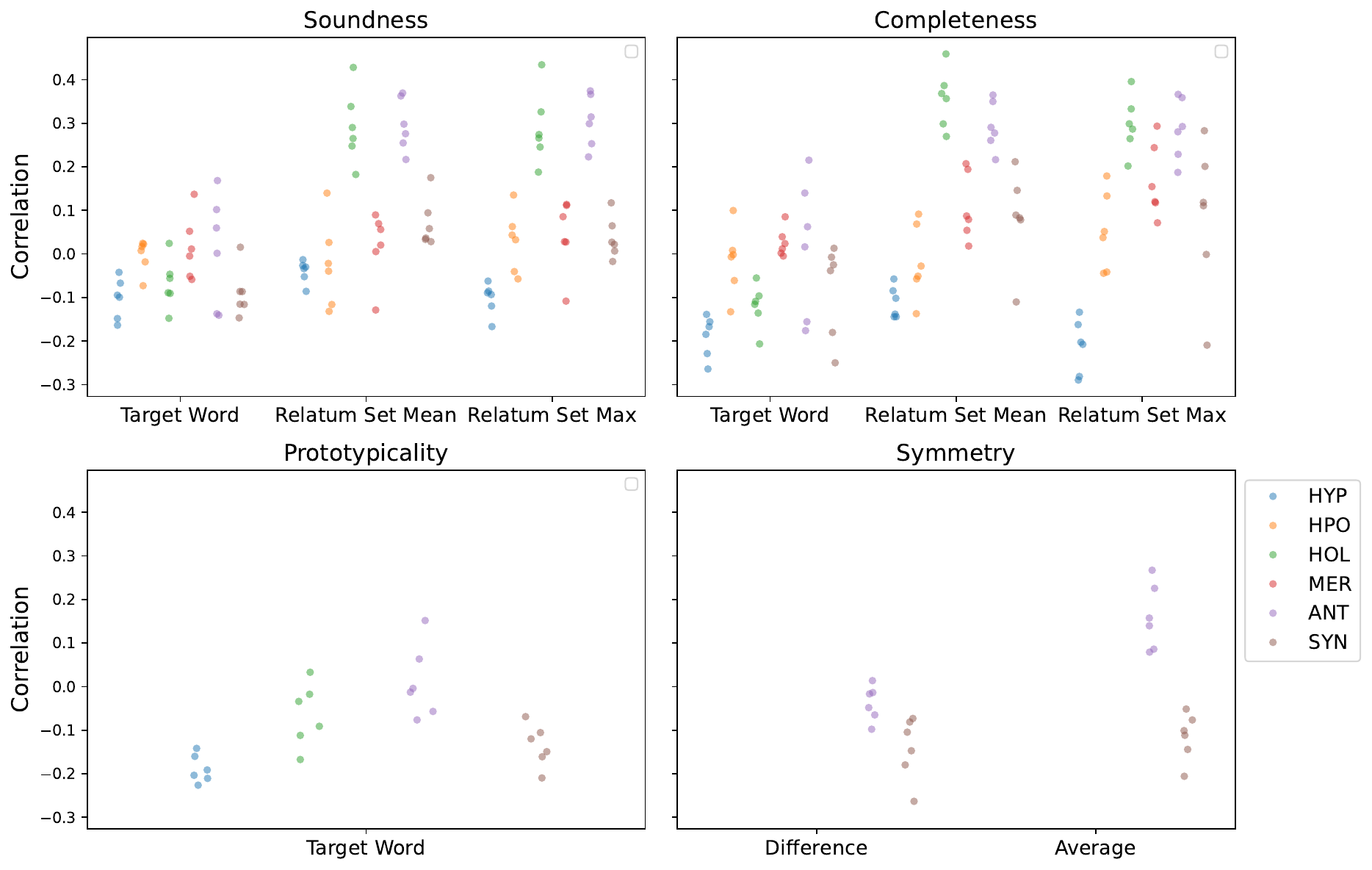}
    \caption{Spearman's $\rho$ between soundness, completeness, asymmetry, symmetry, and prototypicality against word-frequency metrics.}
    \label{fig:corr_sc}
\end{figure}

Figure~\ref{fig:corr_sc} presents the results of all coefficients per metric and relation.

\section{Performance Heteroscedasticity Introduced by Prompts}
\label{app:prompts_hetero}
Prompts are known to influence in spectrum of tasks~\citep{Elazar_2021, Cao_2021}.
We wonder if this finding holds in semantic relation tasks as well.
In order to figure it out, we assess whether prompts introduce performance heteroscedasticity, where performance variances differ significantly across prompts, for agents. 
The heteroscedasticity indicates that an agent's performance originates from populations with varying variances when using different prompts. 
Therefore, observing the heteroscedasticity suggests the influences of prompts for the agent. 

We use Levene's test to determine heteroscedasticity, interpreting the results at a significance level of 0.05.
We target soundness, completeness, and symmetry.
Prototypicality is excluded because not all prompts are used in its calculation~(c.f. Section~\ref{sec:human_pe}).
For each relation and its prompts, we calculate metrics as if each prompt is the only one in the set. Thus, for a given metric and relation, we obtain $N$ sets of results with $N$ prompts.
We apply Levene's test on these $N$ sets.

Results are presented in Table~\ref{tab:prompts_hetero}.
Only models exhibit prompt heteroscedasticity, with the number of such models varying by metrics and relations. 
No sufficiently strong evidence is found supporting the heteroscedasticity in human performance for any metrics and relations.

\begin{table}[htpb]
\caption{Agents, given per metric and relation, present performance heteroscedasticity introduced by prompts. 
The column ``TOTAL'' shows the number of such agents out of 11 agents~(10 models + humans).}
\label{tab:prompts_hetero}
    \begin{tabular}{l|c|l|r}
        \toprule
        Metric & Relation & Agents & TOTAL \\
        \midrule
        \multirow{6}{*}{Soundness}
        & HYP & All models & 6 \\
        & HPO & All models & 6 \\
        & HOL & All models & 6 \\
        & MER & B2, R1, R2, L2 & 4 \\
        & ANT & B1, B2, L1, L2 & 4 \\
        & SYN & B2, R2, L1, L2 & 4 \\
        \midrule
        \multirow{6}{*}{Completeness} 
        & HYP & All models & 6 \\
        & HPO & B1, B2, L1, L2 & 4 \\
        & HOL & All models & 6 \\
        & MER & B2, R1, R2, L2 & 4 \\
        & ANT & B1, B2, L1, L2 & 4 \\
        & SYN & B1, R2, L1, L2 & 4 \\
        \midrule
        \multirow{2}{*}{Symmetry} 
        & ANT & L1, L2 & 2 \\
        & SYN & All models & 6 \\
        \bottomrule
    \end{tabular}
\end{table}

The prompt that achieves the best performance for each relation, metric, and agent is given in the following tables.

\begin{table}[htpb]
\caption{Prompts leading to the best performance per relation and metric for BERT-base.}
\label{tab:best_prompts_B1}
    \begin{tabular}{l|c|l|c}
        \toprule
        Metric & Relation & Prompt & Performance \\
        \midrule
        \multirow{6}{*}{Soundness} 
        & HYP & {\det} {\W} is a subordinate type of {\V} & 0.22\\
        & HPO & {\det} {\W}, such as {\det}  {\V} & 0.22\\
        & HOL & {\det} {\W} is a part of {\det}  {\V} & 0.31\\
        & MER & {\det} {\W} has {\det}  {\V} & 0.13\\
        & ANT & it is not likely to be both {\det} {\W} and {\det}  {\V} & 0.52\\
        & SYN & {\det} {\W} is also known as {\det}  {\V} & 0.06\\
        \midrule
        \multirow{6}{*}{Completeness} 
        & HYP & {\det} {\W} is a subordinate type of {\V} & 0.22\\
        & HPO & {\det} {\W}, such as {\det}  {\V} & 0.16\\
        & HOL & {\det} {\W} is a part of {\det}  {\V} & 0.25\\
        & MER & {\det} {\W} has {\det}  {\V} & 0.13\\
        & ANT & it is not likely to be both {\det} {\W} and {\det}  {\V} & 0.49\\
        & SYN & {\det} {\W} is often referred to as {\det}  {\V} & 0.13\\
        \midrule
        \multirow{4}{*}{Prototypicality} 
        & HYP & {\det} {\W} is a kind of {\det}  {\V} & 0.18\\
        & HOL & {\det} {\W} is a part of {\det}  {\V} & 0.27\\
        & ANT & it is not likely to be both {\det} {\W} and {\det}  {\V} & 0.41\\
        & SYN & {\det} {\W} is often referred to as {\det}  {\V} & 0.14\\
        \midrule
        \multirow{2}{*}{Symmetry} 
        & ANT & it is not likely to be both {\det} {\W} and {\det}  {\V} & 0.61\\
        & SYN & {\det} {\W} is often referred to as {\det}  {\V} & 0.26\\
        \midrule
    \end{tabular}
\end{table}

\begin{table}[htpb]
\caption{Prompts leading to the best performance per relation and metric for BERT-large.}
\label{tab:best_prompts_B2}
    \begin{tabular}{l|c|l|c}
        \toprule
        Metric & Relation & Prompt & Performance \\
        \midrule
        \multirow{6}{*}{Soundness} 
        & HYP & {\det} {\W} is a subordinate type of {\V} & 0.34\\
        & HPO & {\det} {\W}, such as {\det}  {\V} & 0.29\\
        & HOL & {\det} {\W} belongs to parts of {\det}  {\V} & 0.36\\
        & MER & parts of {\det} {\W} include {\det}  {\V} & 0.22\\
        & ANT & it is not likely to be both {\det} {\W} and {\det}  {\V} & 0.65\\
        & SYN & {\det} {\W} is also known as {\det}  {\V} & 0.14\\
        \midrule
        \multirow{6}{*}{Completeness} 
        & HYP & {\det} {\W} is a subordinate type of {\V} & 0.24\\
        & HPO & {\det} {\W}, such as {\det}  {\V} & 0.21\\
        & HOL & {\det} {\W} is a part of {\det}  {\V} & 0.26\\
        & MER & parts of {\det} {\W} include {\det}  {\V} & 0.18\\
        & ANT & it is not likely to be both {\det} {\W} and {\det}  {\V} & 0.62\\
        & SYN & the word {\W} has a similar meaning as the word {\V} & 0.17\\
        \midrule
        \multirow{4}{*}{Prototypicality} 
        & HYP & {\det} {\W} is a subordinate type of {\V} & 0.23\\
        & HOL & {\det} {\W} is a part of {\det}  {\V} & 0.31\\
        & ANT & it is impossible to be both {\det} {\W} and {\det}  {\V} & 0.53\\
        & SYN & {\det} {\W} is also known as {\det}  {\V} & 0.21\\
        \midrule
        \multirow{2}{*}{Symmetry} 
        & ANT & it is impossible to be both {\det} {\W} and {\det}  {\V} & 0.73\\
        & SYN & the word {\W} has a similar meaning as the word {\V} & 0.37\\
        \midrule
    \end{tabular}
\end{table}

\begin{table}[htpb]
\caption{Prompts leading to the best performance per relation and metric for RoBERTa-base.}
\label{tab:best_prompts_R1}
    \begin{tabular}{l|c|l|c}
        \toprule
        Metric & Relation & Prompt & Performance \\
        \midrule
        \multirow{6}{*}{Soundness} 
        & HYP & {\det} {\W} is a subordinate type of {\V} & 0.27\\
        & HPO & {\det} {\W}, such as {\det}  {\V} & 0.24\\
        & HOL & {\det} {\W} is a part of {\det}  {\V} & 0.33\\
        & MER & {\det} {\W} consists of {\det}  {\V} & 0.22\\
        & ANT & {\det} {\W} is the opposite of {\V} & 0.56\\
        & SYN & {\det} {\W} is often referred to as {\det}  {\V} & 0.13\\
        \midrule
        \multirow{6}{*}{Completeness} 
        & HYP & {\det} {\W} is a subordinate type of {\V} & 0.23\\
        & HPO & the word {\W} has a more general sense than the word {\V} & 0.16\\
        & HOL & {\det} {\W} is a component of {\det}  {\V} & 0.25\\
        & MER & components of {\det} {\W} include {\det}  {\V} & 0.18\\
        & ANT & {\det} {\W} is the opposite of {\V} & 0.52\\
        & SYN & the word {\W} has a similar meaning as the word {\V} & 0.13\\
        \midrule
        \multirow{4}{*}{Prototypicality} 
        & HYP & {\det} {\W} is a subordinate type of {\V} & 0.22\\
        & HOL & {\det} {\W} is a part of {\det}  {\V} & 0.28\\
        & ANT & {\det} {\W} is the opposite of {\V} & 0.50\\
        & SYN & {\det} {\W} is also called {\det}  {\V} & 0.18\\
        \midrule
        \multirow{2}{*}{Symmetry} 
        & ANT & {\det} {\W} is the opposite of {\V} & 0.59\\
        & SYN & {\det} {\W} is often referred to as {\det}  {\V} & 0.17\\
        \midrule
    \end{tabular}
\end{table}

\begin{table}[htpb]
\caption{Prompts leading to the best performance per relation and metric for RoBERTa-large.}
\label{tab:best_prompts_R2}
    \begin{tabular}{l|c|l|c}
        \toprule
        Metric & Relation & Prompt & Performance \\
        \midrule
        \multirow{6}{*}{Soundness} 
        & HYP & {\det} {\W} is a type of {\V} & 0.47\\
        & HPO & {\det} {\W}, such as {\det}  {\V} & 0.39\\
        & HOL & {\det} {\W} is a part of {\det}  {\V} & 0.33\\
        & MER & parts of {\det} {\W} include {\det}  {\V} & 0.33\\
        & ANT & if something is {\det} {\W}, then it can not also be {\det}  {\V} & 0.60\\
        & SYN & {\det} {\W} is also called {\det}  {\V} & 0.27\\
        \midrule
        \multirow{6}{*}{Completeness} 
        & HYP & {\det} {\W} is a type of {\V} & 0.26\\
        & HPO & the word {\W} has a more general sense than the word {\V} & 0.20\\
        & HOL & {\det} {\W} is a part of {\det}  {\V} & 0.28\\
        & MER & components of {\det} {\W} include {\det}  {\V} & 0.26\\
        & ANT & if something is {\det} {\W}, then it can not also be {\det}  {\V} & 0.58\\
        & SYN & {\det} {\W} is also called {\det}  {\V} & 0.21\\
        \midrule
        \multirow{4}{*}{Prototypicality} 
        & HYP & {\det} {\W} is a type of {\V} & 0.34\\
        & HOL & {\det} {\W} is a part of {\det}  {\V} & 0.36\\
        & ANT & if something is {\det} {\W}, then it can not also be {\det}  {\V} & 0.52\\
        & SYN & {\det} {\W} is also known as {\det}  {\V} & 0.28\\
        \midrule
        \multirow{2}{*}{Symmetry} 
        & ANT & {\det} {\W} is the opposite of {\V} & 0.73\\
        & SYN & {\det} {\W} is also known as {\det}  {\V} & 0.34\\
        \midrule
    \end{tabular}
\end{table}

\begin{table}[htpb]
\caption{Prompts leading to the best performance per relation and metric for LLaMA-3-8B.}
\label{tab:best_prompts_L1}
    \begin{tabular}{l|c|l|c}
        \toprule
        Metric & Relation & Prompt & Performance \\
        \midrule
        \multirow{6}{*}{Soundness} 
        & HYP & {\det} {\W} is {\det}  {\V} & 0.42\\
        & HPO & the word {\W} has a more general sense than the word {\V} & 0.30\\
        & HOL & {\det} {\W} is a constituent of {\det}  {\V} & 0.39\\
        & MER & constituents of {\det} {\W} include {\det}  {\V} & 0.08\\
        & ANT & the word {\W} has an opposite meaning of the word {\V} & 0.86\\
        & SYN & the word {\W} means nearly the same as the word {\V} & 0.55\\
        \midrule
        \multirow{6}{*}{Completeness} 
        & HYP & {\det} {\W} is a kind of {\det}  {\V} & 0.21\\
        & HPO & the word {\W} has a more general sense than the word {\V} & 0.14\\
        & HOL & {\det} {\W} is a part of {\det}  {\V} & 0.27\\
        & MER & components of {\det} {\W} include {\det}  {\V} & 0.08\\
        & ANT & the word {\W} has an opposite meaning of the word {\V} & 0.82\\
        & SYN & the word {\W} means nearly the same as the word {\V} & 0.38\\
        \midrule
        \multirow{4}{*}{Prototypicality} 
        & HYP & {\det} {\W} is a kind of {\det}  {\V} & 0.29\\
        & HOL & {\det} {\W} is a constituent of {\det}  {\V} & 0.32\\
        & ANT & the word {\W} has an opposite meaning of the word {\V} & 0.68\\
        & SYN & the word {\W} means nearly the same as the word {\V} & 0.39\\
        \midrule
        \multirow{2}{*}{Symmetry} 
        & ANT & the word {\W} has an opposite meaning of the word {\V} & 0.88\\
        & SYN & the word {\W} has a similar meaning as the word {\V} & 0.50\\
        \midrule
    \end{tabular}
\end{table}

\begin{table}[htpb]
\caption{Prompts leading to the best performance per relation and metric for LLaMA-3-70B.}
\label{tab:best_prompts_L2}
    \begin{tabular}{l|c|l|c}
        \toprule
        Metric & Relation & Prompt & Performance \\
        \midrule
        \multirow{6}{*}{Soundness} 
        & HYP & {\det} {\W} is a kind of {\det}  {\V} & 0.46\\
        & HPO & the word {\W} has a more general meaning than the word {\V} & 0.31\\
        & HOL & {\det} {\W} belongs to parts of {\det}  {\V} & 0.45\\
        & MER & constituents of {\det} {\W} include {\det}  {\V} & 0.19\\
        & ANT & the word {\W} has an opposite meaning of the word {\V} & 0.84\\
        & SYN & the word {\W} has a similar meaning as the word {\V} & 0.65\\
        \midrule
        \multirow{6}{*}{Completeness} 
        & HYP & {\det} {\W} is a kind of {\det}  {\V} & 0.22\\
        & HPO & the word {\W} has a more general sense than the word {\V} & 0.15\\
        & HOL & {\det} {\W} is a part of {\det}  {\V} & 0.34\\
        & MER & constituents of {\det} {\W} include {\det}  {\V} & 0.14\\
        & ANT & the word {\W} has an opposite meaning of the word {\V} & 0.81\\
        & SYN & the word {\W} means nearly the same as the word {\V} & 0.44\\
        \midrule
        \multirow{4}{*}{Prototypicality} 
        & HYP & {\det} {\W} is a kind of {\det}  {\V} & 0.31\\
        & HOL & {\det} {\W} belongs to parts of {\det}  {\V} & 0.34\\
        & ANT & the word {\W} has an opposite meaning of the word {\V} & 0.67\\
        & SYN & the word {\W} has a similar meaning as the word {\V} & 0.44\\
        \midrule
        \multirow{2}{*}{Symmetry} 
        & ANT & the word {\W} has an opposite meaning of the word {\V} & 0.92\\
        & SYN & the word {\W} has a similar meaning as the word {\V} & 0.66\\
        \midrule
    \end{tabular}
\end{table}

\end{appendices}

\clearpage
\bibliography{reference}

\end{document}